%% file: main.tex
\documentclass[10pt,twocolumn,letterpaper]{article}

\input{preamble}

\usepackage{cvpr}              

\definecolor{cvprblue}{rgb}{0.21,0.49,0.74}
\usepackage[pagebackref,breaklinks,colorlinks,allcolors=cvprblue]{hyperref}

\usepackage[accsupp]{axessibility}  

\def\paperID{2659} 
\def\confName{CVPR}
\def\confYear{2026}

\title{GenMatter: Perceiving Physical Objects with Generative Matter Models}

\author{Eric Li\textsuperscript{1,*} \quad Arijit Dasgupta\textsuperscript{1} \quad Yoni Friedman\textsuperscript{1} \quad Mathieu Huot\textsuperscript{2}\\[0.3em]
Vikash Mansinghka\textsuperscript{2} \quad Thomas O'Connell\textsuperscript{2} \quad William T. Freeman\textsuperscript{1} \quad Joshua B. Tenenbaum\textsuperscript{1,2}\\[0.5em]
\textsuperscript{1}MIT CSAIL \quad \textsuperscript{2}MIT BCS\\[0.3em]
{\small \textsuperscript{*}Corresponding author: \href{mailto:esli@mit.edu}{\tt esli@mit.edu} \quad Project page: \href{https://esli999.github.io/genmatter/}{\tt esli999.github.io/genmatter}}
}

\begin{document}
\twocolumn[{%
\renewcommand\twocolumn[1][]{#1}%
\maketitle
\begin{center}
    \centering
    \captionsetup{type=figure}
    \includegraphics[width=\linewidth]{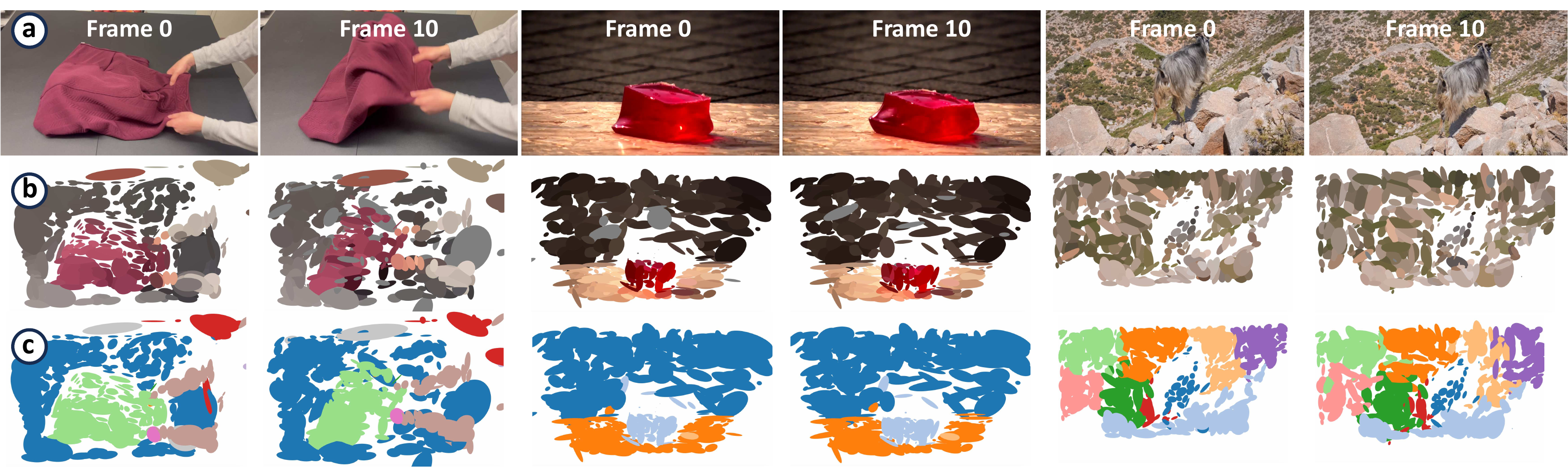}
    \captionof{figure}{\textbf{GenMatter} is a generative model of moving matter. Conditioned on motion and appearance features extracted from RGB video, inference inverts this hierarchical generative model to group observations into \textit{particles} (small Gaussians representing local regions of matter), themselves grouped into \textit{clusters} (coherently and independently moveable physical entities). A hardware-accelerated inference algorithm based on parallelized block Gibbs sampling recovers stable particle motion and groupings. \textbf{(a)} RGB video input. \textbf{(b)} Inferred 3D matter particles shown as colored ellipses, each colored by the average color of its assigned data points. \textbf{(c)} The same particles colored by cluster assignment, revealing independently moving objects.}
    \label{fig:marquee_demo}
\end{center}%
}]
\input{sec/0_abstract}
\input{sec/1_intro}

\input{sec/2_related}

\input{sec/3_method}

\input{sec/4_inference}

\input{sec/5_experiments}

\input{sec/6_discussion}

\section*{Acknowledgements}
This work was supported in part by the Department of the Air Force Artificial Intelligence Accelerator (Cooperative Agreement FA8750-19-2-1000), NSF Award 2019786 (The NSF AI Institute for Artificial Intelligence and Fundamental Interactions), Navy-ONR MURI N00002610, Navy-ONR MURI N00014-22-1-2740, CoCoSys from the Georgia Institute of Technology (Award 2023-JU-3131), the MIT Siegel Family Quest for Intelligence, and the Probabilistic Computing Foundation.

{
    \small
    \bibliographystyle{ieeenat_fullname}
    \bibliography{main}
}

\clearpage
\appendix
\input{sec/appendix}

\end{document}

%% file: preamble.tex
\usepackage{amsmath, amssymb, bm}

\usepackage{booktabs}
\usepackage{algorithm}
\usepackage{algpseudocode}
\usepackage{subcaption}

\usepackage{microtype}



%% file: sec/0_abstract.tex
\begin{abstract}
Human visual perception offers valuable insights for understanding computational principles of motion-based scene interpretation. Humans robustly detect and segment moving entities that constitute independently moveable chunks of matter, whether observing sparse moving dots, textured surfaces, or naturalistic scenes. In contrast, existing computer vision systems lack a unified approach that works across these diverse settings. Inspired by principles of human perception, we propose a generative model that hierarchically groups low-level motion cues and high-level appearance features into particles (small Gaussians representing local matter), and groups particles into clusters capturing coherently and independently moveable physical entities. We develop a hardware-accelerated inference algorithm based on parallelized block Gibbs sampling to recover stable particle motion and groupings. Our model operates on different kinds of inputs (random dots, stylized textures, or naturalistic RGB video), enabling it to work across settings where biological vision succeeds but existing computer vision approaches do not. We validate this unified framework across three domains: on 2D \textit{random dot kinematograms}, our approach captures human object perception including graded uncertainty across ambiguous conditions; on a Gestalt-inspired dataset of camouflaged rotating objects, our approach recovers correct 3D structure from motion and thereby accurate 2D object segmentation; and on naturalistic RGB videos, our model tracks the moving 3D matter that makes up deforming objects, enabling robust object-level scene understanding. This work thus establishes a general framework for motion-based perception grounded in principles of human vision. 
\end{abstract}

%% file: sec/1_intro.tex
\section{Introduction}
\label{sec:intro}

Human vision segments moving objects across diverse settings: random dot kinematograms with minimal form cues~\cite{robert2023disentangling}, camouflaged textured objects perceivable only through motion~\cite{friedman2023benchmarking}, and naturalistic scenes~\cite{nishida2018motion}. No existing computer vision system has this generality. This gap raises a scientific question about the computational principles that enable such broad perceptual capabilities.

\begin{figure*}[tp]
    \centering
    \includegraphics[width=0.9\linewidth]{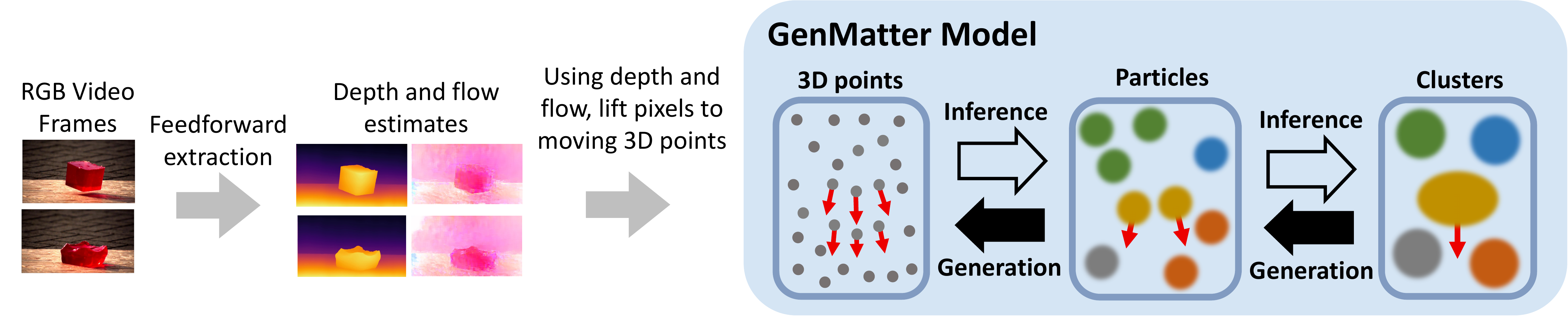}
    \caption{\small \textbf{GenMatter inference pipeline.}
    RGB video is preprocessed (gray arrows) to extract dense depth and optical flow, lifting each pixel to a 3D point tagged with its velocity. The blue box depicts the GenMatter generative model (Sec.~\ref{sec:method}), which represents a scene as a hierarchy of clusters and particles that emit moving 3D points. Black arrows indicate the generative direction (clusters generate particles, which generate 3D points). Hollow arrows indicate the inference algorithm (Sec.~\ref{sec:inference}), which conditions on observed 3D points to infer particle and cluster parameters. Red arrows depict motion at the point, particle, and cluster layers.}
    \label{fig:combined_marquee}
\end{figure*}

We introduce \textbf{GenMatter}, a generative model that segments moving matter across the diverse settings where biological vision succeeds. The model hierarchically groups low-level motion and appearance features into particles (local regions of matter represented as Gaussians), then groups particles into clusters representing coherently and independently moveable entities. The same Bayesian inference algorithm operates across diverse inputs by performing motion feature extraction, which is effective even in abstract textured scenes, and can be supplemented with shape and appearance features for naturalistic video. By jointly inferring particle trajectories and their organization into Spelke objects~\cite{spelke1990principles}, discovering cluster assignments and cluster-level rigid transformations, our approach identifies which particles belong to which moving entities while inducing soft rigidity priors within each entity. Unlike prior methods that impose hand-crafted regularization constraints and avoid explicit object-level grouping, GenMatter captures entities undergoing significant deformation. Figure~\ref{fig:combined_marquee} illustrates the pipeline from RGB frames to inferred particles and clusters.

We evaluate GenMatter across three diverse settings where biological vision robustly detects and groups independently moving matter, yet no single computer vision system succeeds across all three:
\begin{itemize}[leftmargin=*,noitemsep,topsep=0pt]
    \item In random dot kinematogram (RDK) stimuli, where dots follow rigid motion or flicker randomly and correspondences are ambiguous, we test whether motion patterns alone suffice to infer object identity. GenMatter reproduces human perceptual groupings across all difficulty levels and captures graded uncertainty that aligns with intersubject variability.
    \item We evaluate GenMatter on a new dataset of rotating objects with camouflaged textures inspired by Gestalt grouping principles. Human observers readily perceive 3D structure from motion in these stimuli. GenMatter recovers correct 3D structure, with groupings of independently moving matter emerging from probabilistic inference.
    \item On naturalistic RGB videos, GenMatter maintains accurate tracking of the moving 3D matter that makes up deforming objects, matching the performance of supervised trackers without task-specific training.
\end{itemize}

Our evaluation strategy has two complementary objectives. First, we compare GenMatter against standard computer vision approaches across all three settings, demonstrating that both biological vision and GenMatter possess generality that existing computer vision systems lack. Second, within each setting, we provide baselines and ablations to demonstrate specific technical advantages of our structured probabilistic approach: capturing graded perceptual uncertainty, integrating noisy motion signals over time, and generalizing without task-specific training (where supervised baselines may match our performance but lack our cross-domain generality).

We emphasize that GenMatter provides a unifying framework that describes a class of models for motion-based perception. While the class of models defined by our approach requires pretrained feature extractors, the core inference engine requires no task or domain-specific training and fits within a few kilobytes of source code. By recasting perceptual grouping as online probabilistic inference, the structured prior defined by GenMatter's generative model achieves performance comparable to data-driven learning across all studied perception tasks, spanning abstract dot stimuli and naturalistic videos.

%% file: sec/2_related.tex
\section{Related Work}
\label{sec:related}

\textbf{Analysis-by-synthesis}
The analysis-by-synthesis approach posits perception as inference in a structured generative model. Purely model-based systems using probabilistic graphics programs have achieved success in CAPTCHA-breaking, pose estimation, and scene understanding~\cite{mansinghka2013approximate, kulkarni2015picture, gothoskar20213dp3, gothoskar2023bayes3d}, while hybrid approaches combining learned components with structured priors have improved accuracy and efficiency~\cite{eslami2016attend, rempe2021humor, zhou20233d, tewari2023diffusion}. Programming systems supporting inference automation have enabled competitive performance for object detection and tracking~\cite{lew2023smcp3, becker2024probabilistic}.

\noindent
\textbf{Generative models of human perception}
Perception as Bayesian inference in structured generative models captures key aspects of human visual processing, including motion perception, shape recognition, and facial identification~\cite{yildirim2020efficient, hamilton2024seeing, bill2020hierarchical, erdogan2017visual}. This framework extends to physical scene understanding, modeling how humans infer object properties and interactions~\cite{wu2015galileo, battaglia2013simulation, yildirim2024perception, ullman2017mind, wang2024probabilistic, smith2019modeling}. However, computational complexity forces such models to impose restrictive assumptions on texture and geometry. Recent work shows that biologically-inspired motion energy features improve motion segmentation robustness and enable zero-shot generalization to random dot stimuli~\cite{tangemann2024object, sun2025machine}. GenMatter provides a tractable probabilistic model achieving human-like robustness to noise and ambiguity while operating on naturalistic visual input.

\noindent
\textbf{Tracking any point}
The problem of tracking any point was introduced as an intermediate approach between sparse feature tracking and dense optical flow, aiming to capture both long-range and dense motion trajectories~\cite{sand2008particle}. More recent systems have improved tracking performance, producing denser and more robust trajectories across time~\cite{karaev2024cotracker, karaev2024cotracker3, harley2022particle, doersch2023tapir, wang2023tracking}. Extensions to 3D tracking have proceeded along two main directions: some approaches incorporate monocular depth estimation directly into the model~\cite{koppula2024tapvid}, while others infer depth implicitly from learned features over point trajectories~\cite{cho2025seurat}. Further work has used as-rigid-as-possible regularization to improve the quality of point tracking~\cite{xiao2024spatialtracker}. We note that while systems that track any point let users track pre-specified points, they do not offer a way to infer a good point representation. These systems do not pick particle representations adaptively from the scene, nor leverage semantic information for tracking and grouping.

\noindent
\textbf{Motion-based grouping and segmentation}
Contemporary approaches to discovering object structure from motion rely predominantly on learned representations. Motion segmentation methods combine optical flow with appearance priors through diverse training schemes~\cite{dave2019towards, xie2024moving, venkatesh2025discovering}, while foundation models like SAM2~\cite{ravi2024sam} achieve promptable segmentation via large-scale supervision, capturing both whole objects and parts without semantic constraints (the stuff-versus-things distinction~\cite{adelson2001seeing}). Unsupervised methods that learn articulated 3D structure from video~\cite{noguchi2022watch, wan2024template, bao2023object, baekdreamweaver} typically require pretraining, multi-view input, or focus on rigid categories. Methods using internet-scale priors~\cite{li2024learning, li2024dessie} infer static 3D structure but do not aggregate information through multiple frames of a scene. In contrast, GenMatter infers dynamic 3D matter representations of deformable objects via a structured probabilistic prior, providing richer scene understanding than 2D segmentation.

%% file: sec/3_method.tex
\section{Generative Matter Model}
\label{sec:method}

We introduce the Generative Matter Model (\textbf{GenMatter}), a two-level hierarchical generative model for structured motion of deformable matter, defined procedurally in Algorithm~\ref{alg:genParticles}. \textit{Clusters} represent coherent groups, each parameterized by a Gaussian over space and a rigid-body transformation. \textit{Particles} are local Gaussians drawn from clusters that encode spatially localized data points. Data points, each a position-velocity observation constructed from depth and optical flow (Figure~\ref{fig:combined_marquee}), are sampled from a mixture over particles. While cluster transformations encode rigid motion, the particle velocity covariance $\boldsymbol{\Sigma}_\ell^\mathcal{V}$ gives slack to model intra-cluster motion, enabling the model to naturally accommodate both rigid and deformable objects.

\noindent
\textbf{Hierarchical structure}
The algorithm begins by sampling mixture weights $\boldsymbol{\pi}^{\mathcal{H}} \sim \text{Dir}(\boldsymbol{\alpha})$ and $\boldsymbol{\pi}^{\mathcal{B}} \sim \text{Dir}(\boldsymbol{\beta})$ for clusters and particles. Each cluster $k$ is parameterized by a spatial distribution $(\boldsymbol{\mu}_k^\mathcal{H}, \boldsymbol{\Sigma}_k^\mathcal{H})$ and rigid transformation $(\mathbf{R}_k, \mathbf{t}_k)$ drawn from discretized priors suited to small frame-to-frame motions. Each particle $\ell$ is assigned to cluster $k = z_\ell^{\mathcal{H}}$ and samples its spatial mean from $\mathcal{N}(\boldsymbol{\mu}_k^\mathcal{H}, \boldsymbol{\Sigma}_k^\mathcal{H})$, with covariance $\boldsymbol{\Sigma}_\ell^\mathcal{B}$ to explain local matter. Observed points $\mathbf{x}_n$, each a position-velocity pair derived from depth and optical flow as illustrated in Figure~\ref{fig:combined_marquee}, are drawn from particle $\ell = z_n^{\mathcal{B}}$ via $\mathcal{N}(\boldsymbol{\mu}_\ell^\mathcal{B}, \boldsymbol{\Sigma}_\ell^\mathcal{B})$. Optionally, to incorporate image features $\mathbf{f}_n$, we define augmented data points $\tilde{\mathbf{x}}_n = [\mathbf{x}_n; \mathbf{f}_n]$ with $\tilde{\mathbf{x}}_n \sim \mathcal{N}(\boldsymbol{\tilde{\mu}}_\ell, \boldsymbol{\tilde{\Sigma}}_\ell)$. $\mathbf{f}_n$ is assumed to be zero-mean, and $\boldsymbol{\tilde{\Sigma}}_\ell$ is block-diagonal, so the spatial and feature dimensions are assumed to be independent.


\begin{algorithm}[t]
\small
\caption{Generative Particle Model}
\label{alg:genParticles}
\begin{algorithmic}
\State \textbf{Input:} $K, L, N$ (num. clusters, particles, data points)
\State \hspace{1em} Priors: $\boldsymbol{\alpha}, \boldsymbol{\beta}$ (mixture); $\boldsymbol{\mu}^\mathcal{H}, \sigma^2_{\mu^\mathcal{H}}, \boldsymbol{\Psi}^\mathcal{H}, \nu^\mathcal{H}$ (cluster);
\State \hspace{1em} $\boldsymbol{\Psi}^\mathcal{B}, \nu^\mathcal{B}$ (particle matter); $\sigma_V^2, \boldsymbol{\Psi}^\mathcal{V}, \nu^\mathcal{V}$ (velocity)

\State Sample cluster weights: $\boldsymbol{\pi}^{\mathcal{H}} \sim \text{Dir}(\boldsymbol{\alpha})$
\State Sample particle weights: $\boldsymbol{\pi}^{\mathcal{B}} \sim \text{Dir}(\boldsymbol{\beta})$

\For{$k = 1$ to $K$}
    \State Sample cluster covariance: $\boldsymbol{\Sigma}_k^\mathcal{H} \sim \mathcal{W}^{-1}(\boldsymbol{\Psi}^\mathcal{H}, \nu^\mathcal{H})$
    \State Sample cluster mean: $\boldsymbol{\mu}_k^\mathcal{H} \sim \mathcal{N}(\boldsymbol{\mu}^\mathcal{H}, \sigma^2_{\mu^\mathcal{H}} \mathbf{I})$
    \State Sample cluster translation: $\mathbf{t}_k \sim \text{DiscreteNormal}(\mathbf{0}, s^2\mathbf{I})$
    \State Sample cluster rotation: $\mathbf{R}_k \sim \text{DiscreteVMF}(\kappa^{\text{vmf}}, \theta_{\text{max}})$
\EndFor

\For{$\ell = 1$ to $L$}
    \State Sample cluster assignment: $z_\ell^{\mathcal{H}} \sim \text{Cat}(\boldsymbol{\pi}^{\mathcal{H}})$
    \State Let $k = z_\ell^{\mathcal{H}}$
    \State Sample particle covariance: $\boldsymbol{\Sigma}_\ell^\mathcal{B} \sim \mathcal{W}^{-1}(\boldsymbol{\Psi}^\mathcal{B}, \nu^\mathcal{B})$
    \State Sample particle mean: $\boldsymbol{\mu}_\ell^\mathcal{B} \sim \mathcal{N}(\boldsymbol{\mu}_k^\mathcal{H}, \boldsymbol{\Sigma}_k^\mathcal{H})$
    \State Compute cluster-induced velocity:
    \State \hspace{1em} $\bar{\mathbf{v}}_\ell = \mathbf{t}_k + (\mathbf{R}_k - \mathbf{I})(\boldsymbol{\mu}_\ell^\mathcal{B} - \boldsymbol{\mu}_k^\mathcal{H})$
    \State Sample particle velocity mean: $\mathbf{v}_\ell \sim \mathcal{N}(\bar{\mathbf{v}}_\ell, \sigma_V^2 \mathbf{I})$
    \State Sample particle velocity covariance: $\boldsymbol{\Sigma}_\ell^\mathcal{V} \sim \mathcal{W}^{-1}(\boldsymbol{\Psi}^\mathcal{V}, \nu^\mathcal{V})$
\EndFor

\For{$n = 1$ to $N$}
    \State Sample particle assignment: $z_n^{\mathcal{B}} \sim \text{Cat}(\boldsymbol{\pi}^{\mathcal{B}})$
    \State Let $\ell = z_n^{\mathcal{B}}$
    \State Sample data point position: $\mathbf{x}_n \sim \mathcal{N}(\boldsymbol{\mu}_\ell^\mathcal{B}, \boldsymbol{\Sigma}_\ell^\mathcal{B})$
    \State Sample data point velocity: $\mathbf{v}_n \sim \mathcal{N}(\mathbf{v}_\ell, \boldsymbol{\Sigma}_\ell^\mathcal{V})$
\EndFor

\end{algorithmic}
\end{algorithm}

\noindent
\textbf{Velocity model}
The per-particle cluster-induced velocity $\bar{\mathbf{v}}_\ell = \mathbf{t}_k + (\mathbf{R}_k - \mathbf{I})(\boldsymbol{\mu}_\ell^\mathcal{B} - \boldsymbol{\mu}_k^\mathcal{H})$ captures expected motion from the parent cluster's rigid transformation, where $(\mathbf{R}_k - \mathbf{I})$ provides a first-order approximation of rotation about $\boldsymbol{\mu}_k^\mathcal{H}$. The particle velocity mean $\mathbf{v}_\ell \sim \mathcal{N}(\bar{\mathbf{v}}_\ell, \sigma_V^2 \mathbf{I})$ introduces isotropic noise to allow particles to deviate from rigidity, while the covariance $\boldsymbol{\Sigma}_\ell^\mathcal{V} \sim \mathcal{W}^{-1}(\boldsymbol{\Psi}^\mathcal{V}, \nu^\mathcal{V})$ allows data point velocities $\mathbf{v}_n \sim \mathcal{N}(\mathbf{v}_\ell, \boldsymbol{\Sigma}_\ell^\mathcal{V})$ to deviate from particle motion.

\noindent
\textbf{Comparison to ARAP regularization}
To connect our probabilistic approach to optimization-based techniques more common in the literature, we note that a common way to regularize tracking models is using an \emph{as-rigid-as-possible} (ARAP) assumption~\cite{luiten2024dynamic, xiao2024spatialtracker}. At frame $t$, an ARAP regularizer promotes locally rigid motion by penalizing changes in pairwise distances between each point $\mathbf{x}_n^t$ and its neighbors within a fixed radius $r$. In its typical formulation as a loss function, the ARAP is written as:
\[
\mathcal{L}_{\text{ARAP}} = \sum_{d(\mathbf{x}_m^t, \mathbf{x}_n^t) < r}
w_{m,n} \left\| d(\mathbf{x}_m^{t+1}, \mathbf{x}_n^{t+1}) - d(\mathbf{x}_m^t, \mathbf{x}_n^t) \right\|_1,
\]
where $d(\cdot, \cdot)$ is a distance metric. This loss encourages locally-rigid motion within all spheres of radius $r$. $w_{m,n}$ is a weighting given by either a predefined pointwise kernel function $k(\mathbf{x}_m^0,\mathbf{x}_n^0)$ or a learned similarity metric $s(m,n)$ defined on pairs of particle indices.

\noindent
\textbf{Connection to ARAP via small-variance asymptotics}
We can relate certain aspects of GenMatter's particle motion model to ARAP by deriving its small-variance asymptotic limit~\cite{kulkis2012revisiting, broderick2013mad}. Taking $\epsilon/\eta \rightarrow 0$ (where $\epsilon$ and $\eta$ are datapoint-to-particle and particle-to-cluster noise scales) yields a K-means-like objective that alternates between computing centroids and solving for optimal rigid transforms via Procrustes alignment. Letting $\mathbf{x}_n' = \mathbf{R}_k (\mathbf{x}_n - \boldsymbol{\mu}_k^{\mathcal{H}}) + \boldsymbol{\mu}_k^{\mathcal{H}} + \mathbf{t}_k$ denote the predicted position after rigid transformation, the resulting objective is to minimize:
\[
    \mathcal{L}(z_n^{\mathcal{B}}, z_\ell^{\mathcal{H}}) = \sum_n \left\| \mathbf{x}_n + \mathbf{v}_n - \mathbf{x}_n' \right\|_2^2,
    \label{eq:arap}
\]
where $\ell = z_n^\mathcal{B}$. A key distinction is that while ARAP applies rigidity penalties based on fixed distance cutoffs $r$, our approach jointly infers object-centric groupings and rigid motion through hierarchical clustering. By coupling rigidity with probabilistic inference over cluster assignments, the posterior reveals which particles belong to which independently moving entities. However, discrete optimization over assignments is intractable, motivating a hierarchical Bayesian model and probabilistic inference.

\begin{figure}[t]
    \centering
    \includegraphics[width=\linewidth]{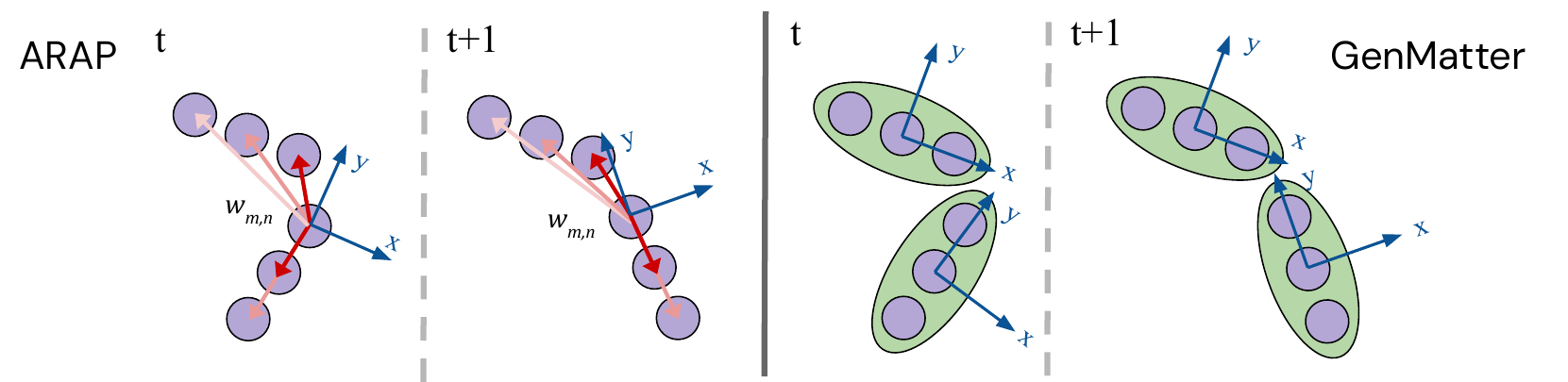}
    \caption{ARAP vs. GenMatter on a two-object scene. ARAP incurs penalties from fixed distance cutoff $r$, while GenMatter infers cluster assignments (green), modeling discontinuous motion.}
    \label{fig:ARAPvGenMatter}
\end{figure}

%% file: sec/4_inference.tex
\section{Inference}
\label{sec:inference}

We perform inference via blocked Gibbs sampling~\cite{jain2004split} that exploits the hierarchical conditional independence structure: variables at each level (data points, particles, clusters) can be updated in parallel given other levels. We leverage conjugate updates (Normal-Inverse-Wishart, Normal-Normal) where possible~\cite{murphy2007conjugate,gelman1995bayesian}. For rigid transforms, we discretize the SE(3) space and enumerate candidates via parallel likelihood evaluation. Our vectorized implementation in the GenJAX probabilistic programming framework~\cite{genjax,genjax_vi,jax2018github} runs efficiently on a single NVIDIA L4 GPU (24GB). We maintain a single-sample posterior approximation, sufficient for high-quality inference in practice. Full derivations are in Appendix~\ref{app:gibbs}.

\noindent
\textbf{Assignment updates}
Datapoints are assigned to particles via a categorical Gibbs conditional combining spatial and velocity likelihoods:
\[
p(z_n^{\mathcal{B}} = \ell \mid \cdot) \propto
\pi_\ell^{\mathcal{B}} \cdot
\mathcal{N}(\mathbf{x}_n \mid \boldsymbol{\mu}_\ell^{\mathcal{B}}, \boldsymbol{\Sigma}_\ell^{\mathcal{B}}) \cdot
\mathcal{N}(\mathbf{v}_n \mid \mathbf{v}_\ell, \boldsymbol{\Sigma}_\ell^{\mathcal{V}}).
\]
Particles are assigned to clusters based on rigid motion fit:
\[
p(z_\ell^{\mathcal{H}} = k \mid \cdot) \propto
\pi_k^{\mathcal{H}} \cdot
\mathcal{N}(\boldsymbol{\mu}_\ell^{\mathcal{B}} \mid \boldsymbol{\mu}_k^{\mathcal{H}}, \boldsymbol{\Sigma}_k^{\mathcal{H}}) \cdot
\mathcal{N}(\mathbf{v}_\ell \mid \bar{\mathbf{v}}_{\ell,k}, \sigma_V^2 \mathbf{I}),
\]
where $\bar{\mathbf{v}}_{\ell,k} = \mathbf{t}_k + (\mathbf{R}_k - \mathbf{I})(\boldsymbol{\mu}_\ell^{\mathcal{B}} - \boldsymbol{\mu}_k^{\mathcal{H}})$ is the velocity derived from cluster $k$'s rigid transformation. Mixture weights are updated via Dirichlet-Categorical conjugacy.

\noindent
\textbf{Parameter updates}
Covariances are updated via Normal-Inverse-Wishart conjugacy using scatter matrices of assigned points. Velocity means $\mathbf{v}_\ell$ are sampled from Gaussian Gibbs conditionals that integrate a rigid motion prior from the parent cluster with observed velocities from assigned datapoints. Particle spatial means $\boldsymbol{\mu}_\ell^{\mathcal{B}}$ are sampled from a Gaussian Gibbs conditional factorized into three terms: (1) a spatial likelihood from the assigned cluster, (2) position likelihoods from assigned datapoints, and (3) a velocity likelihood from rigid motion. 

This velocity likelihood arises because the predicted velocity $\bar{\mathbf{v}}_{\ell,k} = \mathbf{t}_k + (\mathbf{R}_k - \mathbf{I})(\boldsymbol{\mu}_\ell^{\mathcal{B}} - \boldsymbol{\mu}_k^{\mathcal{H}})$ depends linearly on the particle position $\boldsymbol{\mu}_\ell^{\mathcal{B}}$. Since both the prior on $\boldsymbol{\mu}_\ell^{\mathcal{B}}$ and the velocity likelihood $\mathcal{N}(\mathbf{v}_\ell \mid \bar{\mathbf{v}}_{\ell,k}, \sigma_V^2 \mathbf{I})$ are Gaussian with linear dependencies, the Gibbs conditional over $\boldsymbol{\mu}_\ell^{\mathcal{B}}$ remains Gaussian by conjugacy~\cite{gelman2013bayesian}. Cluster means similarly incorporate priors, particle positions, and velocities. Rigid transforms $(\mathbf{R}_k, \mathbf{t}_k)$ are sampled from discretized SE(3), yielding a conjugate Dirichlet-Categorical update.

\noindent
\textbf{Initialization}
We initialize the MCMC chain at frame 0 using hierarchical K-Means clustering, which provides an efficient approximation to burn-in for the Gibbs sampler. We use K-Means++ clustering~\cite{arthur2006k} to initialize particle positions, then run a second K-Means pass to initialize cluster centers. Mixture weights, velocity means, and covariances are set from empirical frequencies and sample statistics. Rigid transforms are initialized via Kabsch alignment~\cite{kabsch1976solution}. Spatial hyperparameters (Inverse-Wishart scale matrices, Gaussian mean priors) encode interpretable priors on cluster and particle size, set from image resolution (see Appendix~\ref{app:init_steps}). Motion hyperparameters are set once per dataset from optical flow statistics. All other hyperparameters remain constant across all videos and datasets.

\noindent
\textbf{Multi-frame tracking}
The two-frame generative model extends to video tracking via sequential MCMC. At frame $t$, particle means are propagated using inferred velocities: $\tilde{\boldsymbol{\mu}}_\ell^{\mathcal{B}, t} = \boldsymbol{\mu}_\ell^{\mathcal{B}, t-1} + \mathbf{v}_\ell^{t-1}$. Since we observe only unordered point clouds $\{\mathbf{x}_n^t\}$ without tracked correspondences, we first assign data points to particles via spatial proximity since $p(z_n^{\mathcal{B}, t} = \ell \mid \mathbf{x}_n^t) \propto \pi_\ell^{\mathcal{B}, t} \cdot \mathcal{N}(\mathbf{x}_n^t \mid \tilde{\boldsymbol{\mu}}_\ell^{\mathcal{B}, t}, \boldsymbol{\Sigma}_\ell^{\mathcal{B}, t})$. After updating particle means from these spatial assignments, subsequent Gibbs sweeps update variables in bottom-up order (data points $\rightarrow$ particles $\rightarrow$ clusters), with assignment distributions now incorporating both position and velocity observations. This ensures cluster inference remains grounded in current observations. Critically, we re-infer cluster assignments and transformations at each frame rather than propagating them, as propagated clusters may incorrectly span articulated parts. This design, inspired by filtering in particle MCMC~\cite{doucet2000sequential,andrieu2010particle}, maintains a tractable posterior while enabling stable tracking.

%% file: sec/5_experiments.tex
\begin{figure*}[t]
    \centering
    \includegraphics[width=\linewidth]{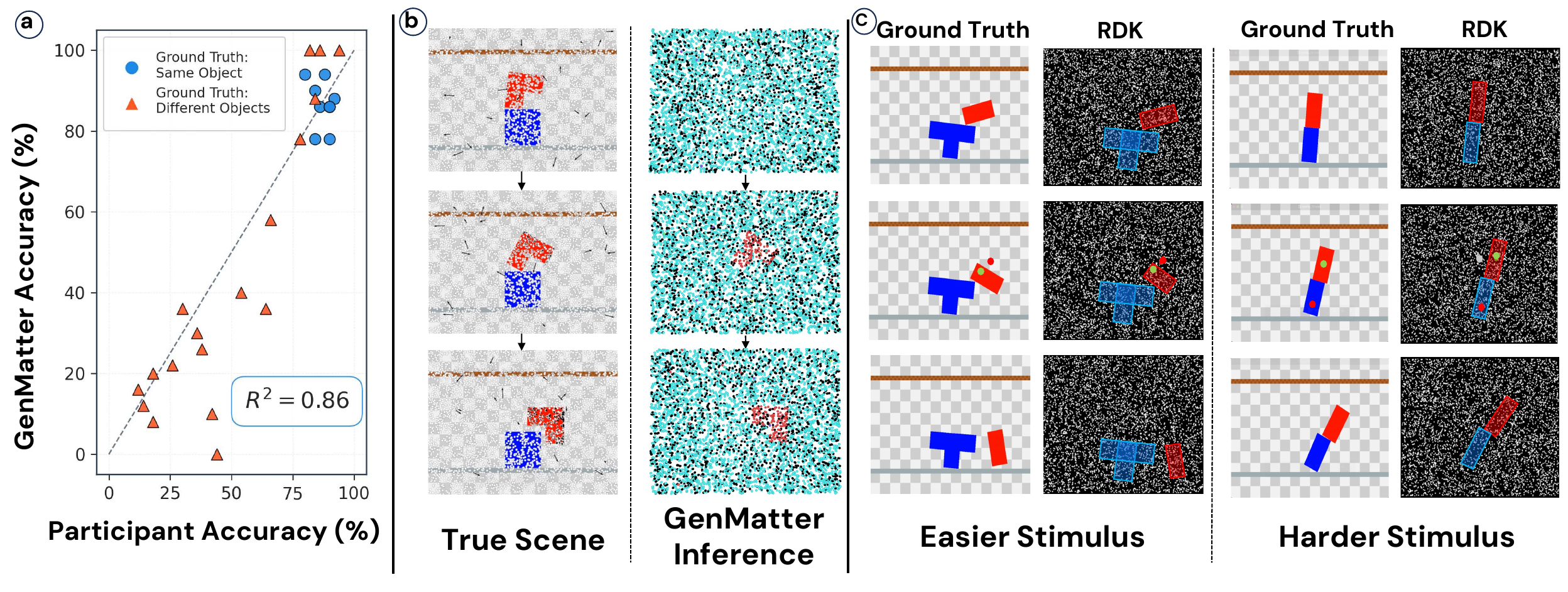}
    \caption{\textbf{GenMatter closely tracks human perceptual judgments on random dot kinematograms.} \textbf{(a)} GenMatter accuracy (\%) vs.\ participant accuracy (\%) across 27 stimuli ($r^2 = 0.86$). Each blue circle represents a same-object stimulus and each orange triangle a different-object stimulus. \textbf{(b)} An internal view of the inferred posterior: red points belong to the moving object, blue points to the background, and black points flickered out of the scene. \textbf{(c)} Fast rotation makes object-background separation easy: both GenMatter (86\%) and humans (90\%) correctly judge the probes as on different objects (left). Slow sliding motion is very challenging: both GenMatter (88\%) and humans (84\%) incorrectly judge the probes as on the same object (right).}
    \label{fig:human_model_comparison}
\end{figure*}

\section{Experiments}
\label{sec:experiments}

We evaluate GenMatter across three settings: 2D random dot kinematograms (RDKs), camouflaged 3D Gestalt stimuli, and naturalistic RGB videos. Figure~\ref{fig:combined_marquee} gives an overview of the complete inference system. We compare against standard computer vision methods and provide within-setting ablations to assess the generality of our probabilistic approach. All CIs are 95\% bootstrap intervals (50{,}000 samples).

\subsection{Human Object Judgments from Motion}
\label{sec:human}

\noindent
\textbf{Stimuli and task} We created 9 unique rigid-body physics scenes using PyMunk~\cite{pymunk2024} and generated 3 RDKs per scene using~\cite{robert2023disentangling}, varying probe dot locations and timings to yield 27 total stimuli. Each stimulus contains object-bound and background dots that either follow rigid motion or flicker randomly, and plays forward then backward to highlight apparent motion. Human participants ($n=150$, 50 per condition) viewed 11 videos each (2 familiarization, 9 experimental) and made binary same-object judgments for red and green probe dots. Participants were recruited from Prolific~\cite{prolific2024} (median duration: 4 minutes; mean age: 37.2; 85 female, 65 male), compensated at local minimum wage, and screened for English fluency, normal color vision, and normal/corrected visual acuity. The study was IRB-approved with fully anonymized data.

\noindent
\textbf{Inference} GenMatter employs a 2D inference pipeline with motion vectors estimated via RANSAC affine transform fitting, ensuring the model operates on image-computable features as humans do when viewing the stimuli. A $k$-nearest neighbor decision policy applied to posterior samples produces binary judgments. GenMatter is run on 50 random seeds per stimulus to match human sample size.

\noindent
\textbf{Results} GenMatter achieves high correlation with human judgments ($r^2 = 0.86$, $t(25) = 12.4$, $p < 0.001$). Figure~\ref{fig:human_model_comparison}a shows variation in human responses across stimuli. Figure~\ref{fig:human_model_comparison}b visualizes GenMatter's internal representation for a single stimulus. The true scene is overlaid with random dots from the kinematogram and their estimated motion vectors. Dots are colored by inferred cluster assignment, with black indicating outliers. These results support latent particle and cluster inference as a computational account of human object perception under motion uncertainty. Figure~\ref{fig:human_model_comparison}c illustrates how some stimuli yield near-unanimous correct responses while others exhibit lower accuracy. GenMatter closely reproduces this graded perceptual uncertainty. The model's ability to reproduce human uncertainty in ambiguous scenes and achieve high accuracy when the scene affords clear grouping establishes it as a valid computational model of human perception across diverse viewing conditions.

\noindent
\textbf{Ablations and baseline} To assess the necessity of hierarchical structure, we evaluate two ablated variants that remove cluster-level variables from the generative model. The fixed particles variant initializes $K=5$ particles with fixed covariance. The adaptive particles variant initializes $K=500$ particles with covariances updated online. Both ablations eliminate the cluster layer, performing inference solely at the particle level. Both variants exhibit substantially degraded correlation with human judgments ($r^2 = 0.35$ and $0.41$, respectively) compared to the full hierarchical model ($r^2 = 0.86$). This demonstrates that the cluster level provides essential structure for capturing human percepts of moving objects. FlowSAM \cite{xie2024moving}, which also relies on motion cues, achieves near-zero correlation with human judgments ($r^2 = 0.04$). This is because random dot kinematograms are a challenging setting where sparse motion extraction is critical, highlighting a fundamental generality gap between human perception and current computer vision systems.

\subsection{Structure from Motion in Gestalt Stimuli}
\label{sec:gestalt}

\begin{table}[t]
\centering
\caption{\textbf{Summary statistics across 140 Gestalt videos.} GenMatter scores higher on mean per-pixel accuracy and Jaccard index. GenMatter is also more consistent across stimuli. Values reported as mean [95\% bootstrap CI].}
\label{tab:camouflage_metrics}
\small
\begin{tabular}{@{}lcc@{}}
\toprule
\textbf{Method} & \textbf{Accuracy} & \textbf{Jaccard} \\
\midrule
SegAnyMo & 0.33 [0.28, 0.37] & 0.26 [0.22, 0.31] \\
FlowSAM & 0.87 [0.85, 0.88] & 0.67 [0.63, 0.70] \\
GenMatter & \textbf{0.94 [0.93, 0.94]} & \textbf{0.72 [0.70, 0.74]} \\
\bottomrule
\end{tabular}
\end{table}

\noindent
\textbf{Stimuli and task} We evaluate GenMatter on a challenging 3D structure-from-motion task where texture provides little information about scene geometry. We created 140 short videos of 3D objects rotating against backgrounds with matched textures. The dataset comprises 20 distinct object geometries, each rendered with 7 different texture patterns matched to their backgrounds. Each 6-frame video sequence is accompanied by ground-truth binary segmentation masks for evaluation. In these camouflaged stimuli, static frames provide minimal segmentation information, yet human observers readily perceive 3D structure from motion~\cite{friedman2023benchmarking}. This design tests whether models rely on per-frame summary statistics or track spatially-localized features across frames. 

\noindent
\textbf{Baselines} We compare GenMatter against two video segmentation methods. SegAnyMo~\cite{huang2025segment} uses point tracking and monocular depth through a learned motion encoder, producing dynamic object masks by grouping tracked points with SAM2. FlowSAM~\cite{xie2024moving} finetunes SAM for optical flow-based video segmentation. GenMatter operates on optical flow and depth. SegAnyMo leverages point tracking (a richer motion representation) and depth, while FlowSAM uses only optical flow. Despite having comparable inputs, GenMatter relies on probabilistic inference rather than learned feed-forward circuits.

\begin{figure*}[t]
    \centering
    \includegraphics[width=0.85\textwidth]{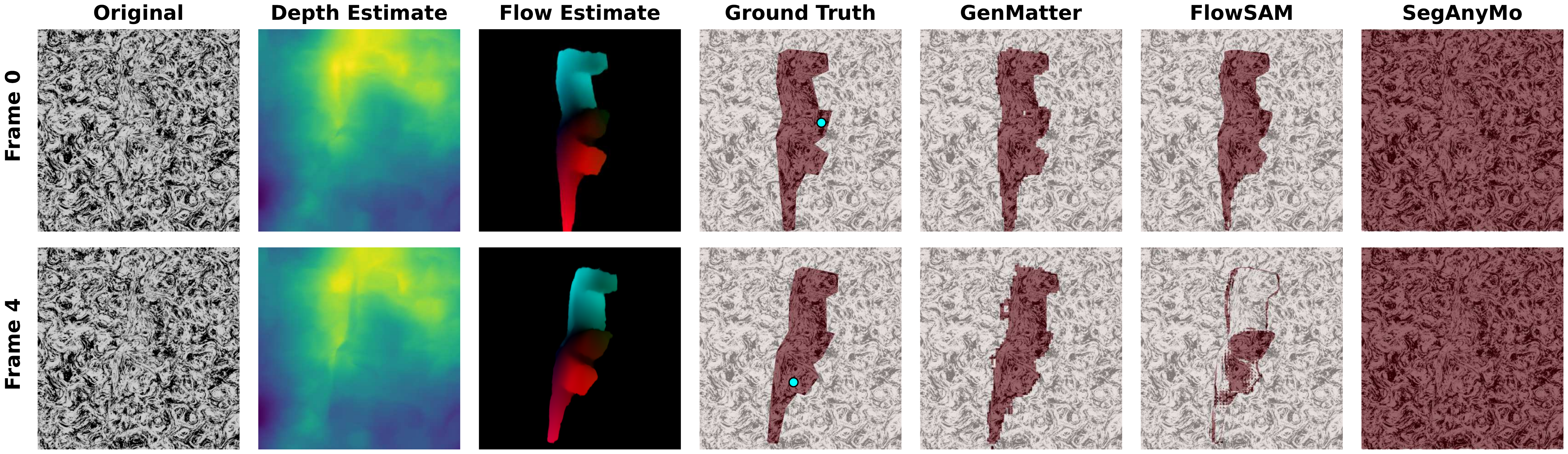}
    \caption{\textbf{Qualitative comparison on camouflaged stimuli.} Probe point segmentation on \texttt{scene\_16}, \texttt{texture\_01}. The depth estimate is uninformative, and the flow estimate shows that on-axis rotation causes opposing motion at top vs.\ bottom (blue vs. red). GenMatter correctly segments the moving object, while FlowSAM segments the initial frame correctly but degrades over time. SegAnyMo fails to detect any object in the scene.}
    \label{fig:gestalt_comparison}
\end{figure*}

\noindent
\textbf{Inference} Our inputs to GenMatter come from optical flow and monocular depth, via RAFT and VideoDepthAnything~\cite{teed2020raft, chen2025video}. All methods are evaluated at $96 \times 96$ resolution. Each video provides 5 flow frames. GenMatter obtains 500 posterior samples per frame via Gibbs sampling to approximate the posterior distribution, from which we extract the maximum a posteriori (MAP) estimate for evaluation.

\noindent
\textbf{Evaluation protocol} Following the probe-point methodology established in the RDK experiments, we evaluate segmentation by sampling query points rather than densely evaluating all pixels. Dense evaluation on all ground-truth pixels requires establishing correspondence between predicted and ground-truth segments, which can implicitly assume a fixed number of objects. Since motion-based discovery aims to identify moving objects without prior knowledge of object count or identity, we adopt a probe-point sampling approach, visualized in Figure~\ref{fig:gestalt_comparison}. For each frame, we sample 100 probe points uniformly from the ground-truth object region $\mathcal{O}$. For each probe location $\mathbf{p}_i$, we identify the predicted segment $S(\mathbf{p}_i)$ containing it and compute pixel-wise accuracy of $S(\mathbf{p}_i)$ against ground truth. Aggregating over probes yields a Monte Carlo estimate of the probabilistic segmentation, given by $p_{\text{pred}}(\text{object} \mid \mathbf{x}) \approx \frac{1}{N} \sum_{i=1}^{N} \mathbb{I}[\mathbf{x} \in S(\mathbf{p}_i)]$ where $\mathbf{p}_i \sim \text{Uniform}(\mathcal{O})$. For each probe point $\mathbf{p}_i$, we compute the Jaccard index $J(S(\mathbf{p}_i), \mathcal{O})$ between its predicted segment $S(\mathbf{p}_i)$ and the ground truth object region $\mathcal{O}$ using the standard IoU formulation. Frame-level accuracy is computed as $\mathbb{E}_{\mathbf{p} \sim \text{Uniform}(\mathcal{O})}[J(S(\mathbf{p}), \mathcal{O})]$, approximated via Monte Carlo averaging over probe samples. Per-video accuracy is then obtained by averaging across all frames in each sequence.

\noindent
\textbf{Results} GenMatter outperforms both supervised baselines, detailed in Table~\ref{tab:camouflage_metrics}, achieving higher Jaccard ($J$) on 111/140 videos against FlowSAM and 133/140 against SegAnyMo ($p < 1 \times 10^{-6}$, paired $t$-test). GenMatter's advantage is consistent across all texture patterns. Figure~\ref{fig:gestalt_comparison} visualizes these performance differences through segmentation overlays, showing that GenMatter assigns concentrated probability mass to contiguous matter regions while FlowSAM produces predictions that degrade over time. On the example shown, SegAnyMo is unable to detect the object ($J = 0.15$), while FlowSAM achieves moderate performance ($J = 0.70$), both falling substantially below GenMatter ($J = 0.92$). Thus, GenMatter's projected 3D representation can outperform learned 2D segmentation on motion-dominated grouping. Monocular depth estimates are sometimes unreliable on camouflaged textures. Ablating depth as a model input entirely still yields accuracy $= 0.89$, slightly above FlowSAM ($0.87$), confirming that optical flow drives performance on this benchmark.

\subsection{3D Particle Representations from RGB Video}
\label{sec:cv}

\noindent
\textbf{Task and motivation} We evaluate whether probabilistic inference in the GenMatter model enables robust particle-based matter representations on videos in TAP-Vid-DAVIS~\cite{doersch2023tapir}. While GenMatter infers full 3D particle representations, we project to 2D tracks for comparison against CoTracker3~\cite{karaev2024cotracker3}, a supervised baseline with the same input-output specification: RGB video $\rightarrow$ particle tracks.

\noindent
\textbf{Model} In this setting, we condition GenMatter on monocular depth~\cite{chen2025video}, optical flow~\cite{teed2020raft}, and DINO~\cite{oquab2023dinov2} features. Each particle maintains a learned appearance vector in DINO space, evolving through Gibbs updates conditioned on cluster assignments. Shared cluster assignments couple appearance to spatial structure, while cluster-level rigid transformations constrain particle motions. Particles and clusters are initialized at frame 0 using SAM2~\cite{ravi2024sam} masks to provide initial proposals for spatial grouping, which are then propagated forward via approximate Bayesian filtering. We use GenMatter with 500 particles with 9 clusters.

\noindent
\textbf{Baselines and metrics} We compare against CoTracker3~\cite{karaev2024cotracker3}, a transformer-based point tracker supervised with simulation data. For consistency across methods, we also initialize CoTracker3 with 500 points.  We note that CoTracker3 tracks individual pixel locations, while GenMatter's particles represent Gaussian regions of matter with spatial extent, and observed data is probabilistically attributed to these particles. We evaluate our model on a projection to 2D tracking rather than 3D particle ground truth due to the scarcity of deformable naturalistic datasets with 3D ground truth annotations. TAP-Vid-DAVIS provides high-fidelity 2D segmentation masks, enabling comparison through projection of GenMatter's 3D particles to 2D. To enable comparison between these fundamentally different representations, we adopt matter-weighted Jaccard $J_m = \sum_i w_i f_i / (\sum_i w_i f_i + \sum_j w_j (1-f_j))$, where $w_i = n_i \pi_i$ weights particle $i$ by spatial extent ($n_i$ pixels) and mixture probability ($\pi_i$), and $f_i$ is fractional overlap with ground truth. This metric accounts for graded uncertainty in GenMatter's probabilistic matter representation. It reduces to standard Jaccard when particles have uniform weights, which we assume for CoTracker3. 

\begin{figure}[t]
    \centering
    \includegraphics[width=\linewidth]{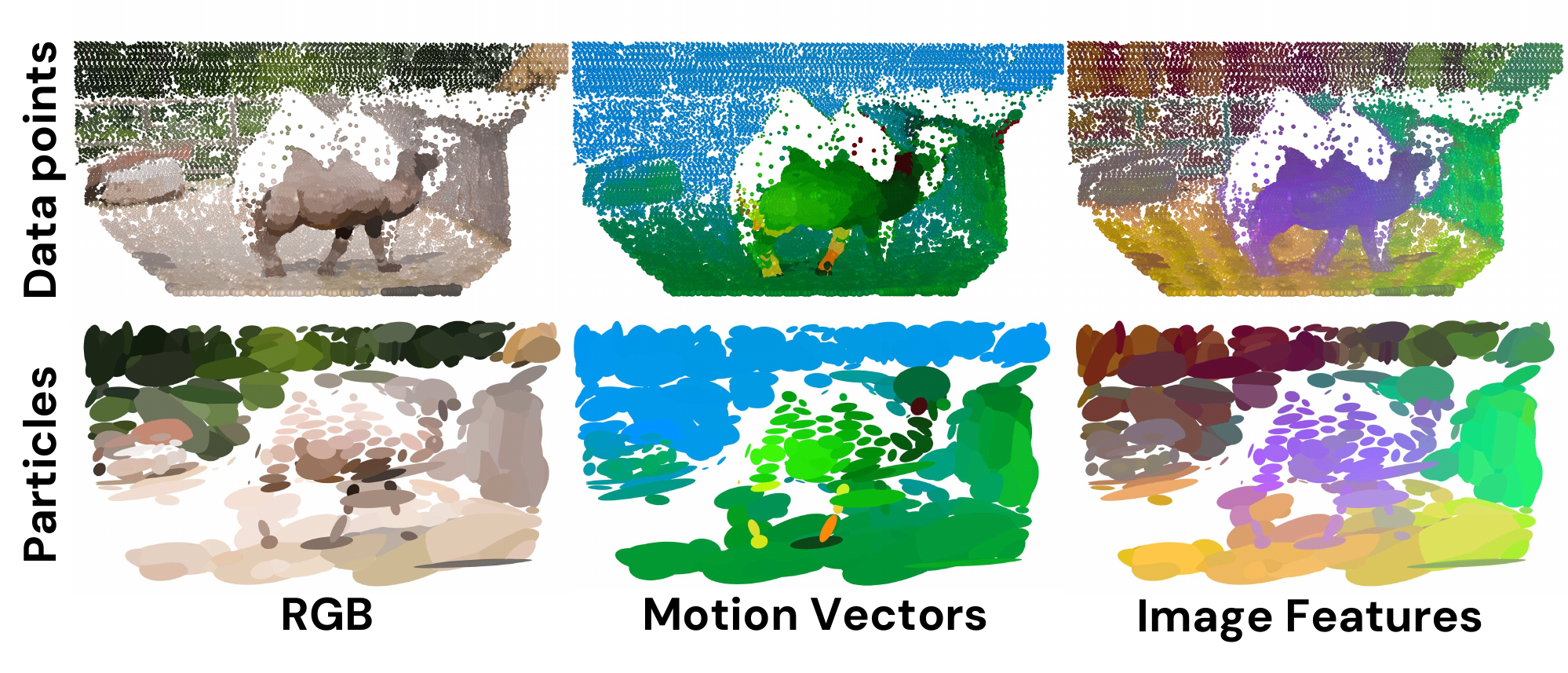}
    \caption{\textbf{Per-point particle assignment visualization.} Each data point is colored by its assigned particle's RGB color (left), motion direction (middle), and appearance features (right). Gaussian particles are shown in the second row. Distinct patterns across motion and appearance demonstrate that GenMatter integrates complementary information sources for faithful matter representation.}
    \label{fig:camel_particle_viz}
\end{figure}

\noindent
\textbf{Results} GenMatter achieves matter-weighted Jaccard of 0.79, matching CoTracker3 (0.78) without task-specific pretraining, shown in Table~\ref{tab:tapvid_tracking}. Unlike learned trackers, GenMatter's hierarchical structure enables explicit integration of SAM2 segmentation proposals into spatial clustering. However, ground-truth initialization degrades GenMatter performance to 0.77. This degradation occurs because ground-truth initialization at frame 0 imposes hard geometric constraints that do not always align with noisy flow and depth estimates. Additionally, ground-truth foreground masks provide only a single binary segmentation without spatial decomposition of background regions, limiting initial clustering of non-object areas. In contrast, SAM2's multi-segment decomposition helps GenMatter model both object and non-object regions effectively. Ablating cluster-level variables degrades performance substantially ($J_m$ from 0.79 to 0.69), confirming that hierarchical structure is essential. Ablating depth as a model input similarly degrades performance ($J_m = 0.69$ [0.61, 0.77] with SAM init, $0.66$ [0.58, 0.73] with GT init), confirming that 3D structure contributes meaningfully beyond motion alone. Figure~\ref{fig:camel_particle_viz} visualizes the particle assignments in an example video. Each pixel's assigned particle is shown by RGB color (left), velocity direction (middle, normalized and color-coded), and DINO features (right, first 3 PCA dimensions). This demonstrates how GenMatter's hierarchical inference integrates position, motion, and appearance into a unified particle representation, enabling structured scene decomposition without task-specific training.

\begin{table}[t]
\centering
\caption{\textbf{Tracking performance on TAP-Vid DAVIS.} GenMatter matches CoTracker3 without task-specific pre-training. Using GT segmentation mask instead of SAM for initialization decreases GenMatter performance but does not affect CoTracker3. Values reported as mean [95\% bootstrap CI].}
\label{tab:tapvid_tracking}
\small
\begin{tabular}{@{}l@{\hspace{4pt}}c@{\hspace{4pt}}c@{\hspace{4pt}}c@{}}
\toprule
\textbf{Metric} & \textbf{CoTracker3} & \textbf{GenMatter} & \textbf{GenMatter (abl.)}\\
\midrule
$J_m$ (SAM) & 0.78 [0.69, 0.87] & \textbf{0.79 [0.73, 0.84]} & 0.69 [0.61, 0.77] \\
$J_m$ (GT) & \textbf{0.78 [0.69, 0.87]} & 0.77 [0.73, 0.84] & 0.68 [0.58, 0.73] \\
\bottomrule
\end{tabular}
\end{table}

\noindent
\textbf{Compute-accuracy tradeoffs} GenMatter's hierarchy enables runtime computational control through data point subsampling at each frame. Since data points occupy the lowest hierarchical level, subsampling them directly reduces the number of latent variables without changing model architecture, in contrast to learned approaches with fixed computation circuits. We evaluate subsampling rates from 1/8 to 1/512 of available data points. Across all rates from 1/8 to 1/128, GenMatter incurs no statistically significant performance loss ($J_m = 0.76$--$0.79$) relative to full resolution ($J_m = 0.76$ [0.71, 0.82]), while running up to $12\times$ faster (9.8 FPS at 1/128 vs.\ 0.80 FPS). Accuracy saturates beyond moderate subsampling because upsampled features provide no additional information once sampling density exceeds the resolution of the lowest-resolution feature (DINO). Performance degrades substantially only at extreme subsampling ($J_m = 0.56$ [0.46, 0.65] at 1/512), where too few observations remain to support reliable inference. This range of operating speeds and accuracies demonstrates flexibility unavailable to end-to-end learned architectures.

%% file: sec/6_discussion.tex
\section{Discussion}
\label{sec:discussion}

\noindent
\textbf{Limitations}
Our model currently represents physical matter without explicit dynamics. A natural extension would be incorporating physics-based dynamics, enabling a joint matter and dynamics model akin to a game engine. Such a model would provide forward prediction of object positions and motion, which would be particularly beneficial for tracking through complete occlusion where GenMatter's current performance degrades due to limited mechanisms for reinitializing or reidentifying completely hidden matter. Our implementation uses a fixed particle count $L$, which limits adaptation to scale changes and objects entering or exiting the scene. This is addressable through dynamic particle allocation strategies from Bayesian nonparametrics.

We provide a general set of evaluation methods across diverse visual perception tasks, though more thorough benchmarking would better assess the quality of both our model and the baselines. The RDK and 3D Gestalt experiments use binary response formats that match typical human experimental protocols, but these collapse posterior distributions into discrete judgments. Extending evaluation to capture per-participant posteriors would enable richer characterization of scene belief states. For particle tracking, we use 2D masks as a proxy for 3D tracking given the current lack of large-scale datasets with ground-truth 3D annotations for deformable objects in naturalistic settings. Developing such datasets through real-world annotation and synthetic rendering would enable more rigorous benchmarks for evaluating 3D matter representations. 

\noindent
\textbf{Broader implications}
GenMatter demonstrates that hierarchical probabilistic inference of particle-based representations provides a unified computational account of motion-based segmentation across minimal dot stimuli, camouflaged objects, and naturalistic scenes. This generality arises from structured priors encoding rigid motion and spatial grouping, rather than hand-crafted regularizers, suggesting that structured probabilistic inference provides essential inductive biases for compositional scene understanding. In doing so, GenMatter bridges scientific models of human visual perception and engineered computer vision systems.

%% file: sec/appendix.tex
\section{Overview of Supplementary Materials}

This supplementary material provides technical details for the main paper. We provide the following supplementary derivations, experimental details, and quantitative results in this document:

\begin{itemize}
    \item Section~\ref{app:SVA}: Derivation of clustering algorithm from small-variance asymptotics
    \item Section~\ref{app:gibbs}: Full derivation of each step in the GenMatter Gibbs sampler
    \item Section~\ref{app:feature_model}: Feature-augmented variant and inference algorithm modifications
    \item Section~\ref{app:human}: Human RDK psychophysics experiment
    \item Section~\ref{app:gestalt}: Gestalt structure-from-motion experiment
    \item Section~\ref{app:3d_first_frames}: Details for all 3D RGB experiments, including first frame visualizations, quantitative results on deformable RGB videos, and technical details about the TAP-Vid-DAVIS benchmark. These videos are meant to capture our model's ability to explain deformable matter in a wide variety of settings.
\end{itemize}



\section{Clustering Algorithm from Small-Variance Asymptotics}
\label{app:SVA}

We present the technical details of the SVA clustering algorithm and recover a rigid group-centric loss function along with an iterative procedure that minimizes it.

\paragraph{Deriving GenMatter as a Clustering Algorithm}

\begin{figure*}[t]
\centering
\begin{minipage}{0.85\textwidth}
\begin{algorithm}[H]
\caption{Clustering Algorithm for GenMatter via Small-Variance Asymptotics}
\begin{algorithmic}[1]
\State \textbf{Input:}
\State \hspace{1em} Number of clusters and particles $K, L$
\State \hspace{1em} Data point positions \(\{\mathbf{x}_n, \mathbf{v}_n\}_{n=1}^N\),
\State \textbf{Initialize:} Assign data points to particles \(z_n^\mathcal{B}\), particles to clusters \(z_\ell^\mathcal{H}\)
\Repeat
    \For{each particle \(\ell = 1, \dots, L\)}
        \State Compute particle mean: \(\boldsymbol{\mu}_\ell^\mathcal{B} \gets \frac{1}{|\mathcal{B}_\ell|} \sum_{n: z_n^\mathcal{B} = \ell} \mathbf{x}_n\)
    \EndFor
    \For{each cluster \(k = 1, \dots, K\)}
        \State Compute cluster mean: \(\boldsymbol{\mu}_k^\mathcal{H} \gets \frac{1}{|\mathcal{H}_k|} \sum_{\ell: z_\ell^\mathcal{H} = k} \boldsymbol{\mu}_\ell^\mathcal{B}\)
        \State Collect point pairs \((\mathbf{x}_n, \mathbf{v}_n)\) assigned to cluster \(k\)
        \State Center points: \(\mathbf{p}_n \gets \mathbf{x}_n - \boldsymbol{\mu}_k^\mathcal{H}\), \(\mathbf{q}_n \gets \mathbf{x}_n + \mathbf{v}_n - \boldsymbol{\mu}_k^\mathcal{H}\)
        \State Compute cross-covariance: \(\mathbf{S}_k \gets \sum_n \mathbf{q}_n \mathbf{p}_n^\top\)
        \State Compute SVD: \(\mathbf{S}_k = \mathbf{U}_k \Sigma_k \mathbf{V}_k^\top\)
        \State Set rotation: \(\mathbf{R}_k \gets \mathbf{U}_k \mathbf{V}_k^\top\)
        \State Set translation: $\mathbf{t}_k = \bar{\mathbf{q}} - \mathbf{R}_k \bar{\mathbf{p}}$
    \EndFor
    \For{each data point \(n = 1, \dots, N\)}
        \State Compute motion loss: \(\mathcal{L}_n(z_n^{\mathcal{B}}, z_\ell^{\mathcal{H}}) \gets \Big\| \mathbf{x}_n + \mathbf{v}_n - \Big( \mathbf{R}_{z_\ell^{\mathcal{H}}} \left( \mathbf{x}_n - \boldsymbol{\mu}_{z_\ell^{\mathcal{H}}}^{\mathcal{H}} \right) + \boldsymbol{\mu}_{z_\ell^{\mathcal{H}}}^{\mathcal{H}} + \mathbf{t}_{z_\ell^{\mathcal{H}}} \Big) \Big\|^2\)
        \State Update \(z_n^\mathcal{B}\), \(z_\ell^\mathcal{H}\) to minimize \(\sum_n \mathcal{L}_n(z_n^{\mathcal{B}}, z_\ell^{\mathcal{H}})\)
    \EndFor
\Until{assignments converge or objective does not decrease}
\end{algorithmic}
\end{algorithm}
\end{minipage}
\end{figure*}

\paragraph{$\boldsymbol{\mu}_\ell^\mathcal{B}$ update:}
In the model, $\boldsymbol{\mu}_\ell^\mathcal{B} \sim \mathcal{N}(\boldsymbol{\mu}_k^\mathcal{H}, \boldsymbol{\Sigma}_k^\mathcal{H})$ and $\mathbf{x}_n^t \sim \mathcal{N}(\boldsymbol{\mu}_\ell^\mathcal{B}, \boldsymbol{\Sigma}_\ell^\mathcal{B})$. Assume $\boldsymbol{\Sigma}_\ell^\mathcal{B} = \epsilon \mathbf{I}$ and $\boldsymbol{\Sigma}_k^\mathcal{H} = \eta \mathbf{I}$, where $\epsilon/\eta \rightarrow0$. The negative log-conditional of $\boldsymbol{\mu}_\ell^\mathcal{B}$ is:

\[
\mathcal{L}(\boldsymbol{\mu}_\ell^\mathcal{B}) \propto \frac{1}{\epsilon} \sum_{n \in \mathcal{B}_\ell} \left\| \mathbf{x}_n^t - \boldsymbol{\mu}_\ell^\mathcal{B} \right\|^2 + \frac{1}{\eta} \left\| \boldsymbol{\mu}_\ell^\mathcal{B} - \boldsymbol{\mu}_k^\mathcal{H} \right\|^2
\]
As $\epsilon/\eta \rightarrow0$, the first term dominates the posterior, so the minimizer of the loss is:
\[
\boldsymbol{\mu}_\ell^\mathcal{B} = \arg\min_{\boldsymbol{\mu}} \sum_{n \in \mathcal{B}_\ell} \| \mathbf{x}_n^t - \boldsymbol{\mu} \|^2  = \frac{1}{|\mathcal{B}_\ell|} \sum_{n \in \mathcal{B}_\ell} \mathbf{x}_n^t.
\]

\paragraph{$\boldsymbol{\mu}_k^\mathcal{H}$ update:}

Assuming $\boldsymbol{\Sigma}_k^\mathcal{H} = \eta \mathbf{I}$ and $\epsilon/\sigma^2_{\mu^\mathcal{H}} \rightarrow0$, the negative log-conditional of $\boldsymbol{\mu}_k^\mathcal{H}$ can be approximated by:

\[
\mathcal{L}(\boldsymbol{\mu}_k^{\mathcal{H}}) \propto \sum_{\ell \in \mathcal{H}_k} \left\| \boldsymbol{\mu}_\ell^{\mathcal{B}} - \boldsymbol{\mu}_k^{\mathcal{H}} \right\|^2
\]
where the minimizer is:

\[
\boldsymbol{\mu}_k^{\mathcal{H}} = \arg\min_\mu \sum_{n \in \mathcal{B}_\ell} \| \boldsymbol{\mu}_\ell^{\mathcal{B}} - \mu \|^2 = \frac{1}{|\mathcal{H}_k|} \sum_{\ell \in \mathcal{H}_k} \boldsymbol{\mu}_\ell^{\mathcal{B}}
\]

\paragraph{$\mathbf{R}_k, \mathbf{t}_k$ update:} We restrict $n$ in this step to only index points that are assigned to cluster $k$. Taking the limit of all noise terms $\sigma \rightarrow 0$ collapses out the dependence on $\boldsymbol{\mu}_\ell^{\mathcal{B}}$ and gives a deterministic motion model that only depends on the relative position of $\mathbf{x}_n$ with respect to $\boldsymbol{\mu}_k^{\mathcal{H}}$. Noting that we also collapse out $\boldsymbol{\Sigma}_\ell^\mathcal{V}$ and $\sigma_V^2$, algebraic manipulation gives that the negative log-conditional of $\mathbf{R}_k, \mathbf{t}_k$ is, with $\mathbf{p}_n = \mathbf{x}_n - \boldsymbol{\mu}_k^{\mathcal{H}}$:

\[
\mathcal{L}_k(\mathbf{R}_k, \mathbf{t}_k) = \sum_{n} \Big\| \mathbf{x}_n + \mathbf{v}_n - \left( \mathbf{R}_k \mathbf{p}_n + \boldsymbol{\mu}_k^{\mathcal{H}} + \mathbf{t}_k \right) \Big\|^2
\]
Letting $\mathbf{q}_n = \mathbf{x}_n + \mathbf{v}_n - \boldsymbol{\mu}_k^{\mathcal{H}}$, the loss becomes:
\[
\mathcal{L}_k(\mathbf{R}_k, \mathbf{t}_k) =
\sum_{n} \left\| \mathbf{q}_n - (\mathbf{R}_k \mathbf{p}_n + \mathbf{t}_k) \right\|^2.
\]
This expression corresponds to the orthogonal Procrustes problem, which has a standard solution. We first define $\bar{\mathbf{p}} = \frac{1}{N} \sum_n \mathbf{p}_n$ and $\bar{\mathbf{q}} = \frac{1}{N} \sum_n \mathbf{q}_n$ and compute $\tilde{\mathbf{p}}_n = \mathbf{p}_n - \bar{\mathbf{p}}$ and $\tilde{\mathbf{q}}_n = \mathbf{q}_n - \bar{\mathbf{q}}$. We then compute the cross-covariance matrix $\mathbf{S}_k = \sum_{n} \tilde{\mathbf{q}}_n \tilde{\mathbf{p}}_n^\top$ and its SVD $\mathbf{S}_k = \mathbf{U}_k \boldsymbol{\Sigma}_k \mathbf{V}_k^\top$. The optimal rotation is $\mathbf{R}_k = \mathbf{U}_k \mathbf{V}_k^\top$ and the optimal translation is $\mathbf{t}_k = \bar{\mathbf{q}} - \mathbf{R}_k \bar{\mathbf{p}}$.

\paragraph{\(z_n^\mathcal{B}\), \(z_\ell^\mathcal{H}\) update:}

Small variance analysis gives the same form of the objective function as the previous step. The negative log-conditional for \(z_n^\mathcal{B}\), \(z_\ell^\mathcal{H}\) then becomes, with \(\mathbf{p}_n = \mathbf{x}_n - \boldsymbol{\mu}_{z_\ell^{\mathcal{H}}}^{\mathcal{H}}\):

\[
\mathcal{L}(z_n^{\mathcal{B}}, z_\ell^{\mathcal{H}}) = \sum_n \left\| \mathbf{x}_n + \mathbf{v}_n - \left( \mathbf{R}_{z_\ell^{\mathcal{H}}} \mathbf{p}_n + \boldsymbol{\mu}_{z_\ell^{\mathcal{H}}}^{\mathcal{H}} + \mathbf{t}_{z_\ell^{\mathcal{H}}} \right) \right\|^2
\]
This is a discrete combinatorial optimization problem that involves searching through particle and cluster assignments.

\section{Blocked Gibbs Sampling}
\label{app:gibbs}

We describe the Gibbs sampling approach in greater detail than in the main text. We first independently describe each blocked Gibbs step in Appendix~\ref{app:gibbs_steps}. Then, we describe the procedure of these steps used for initialization in Appendix~\ref{app:gibbs_init} and tracking in Appendix~\ref{app:gibbs_tracking}.

\subsection{Gibbs Update Steps}
\label{app:gibbs_steps}

There are twelve variables of interest, separated at different hierarchical levels as shown:

\begin{enumerate}
  \item Cluster-level variables:
  \[
  \{\boldsymbol{\mu}_k^{\mathcal{H}},\ \boldsymbol{\Sigma}_k^{\mathcal{H}},\ \mathbf{R}_k,\ \mathbf{t}_k,\ \pi_k^{\mathcal{H}}\}_{k=1}^K
  \]
  \item Particle-level variables:
  \[
  \{\boldsymbol{\mu}_\ell^{\mathcal{B}},\ \boldsymbol{\Sigma}_\ell^{\mathcal{B}},\ \mathbf{v}_\ell,\ \boldsymbol{\Sigma}_\ell^{\mathcal{V}},\ z_\ell^{\mathcal{H}},\ \pi_\ell^{\mathcal{B}}\}_{\ell=1}^L
  \]
  \item Data point-level variables:
  \[
  \{z_n^{\mathcal{B}}\}_{n=1}^N
  \]
\end{enumerate}
For each of these variables, we independently describe each of the Gibbs updates.

\subsubsection{Data point-to-Particle Assignments (\texorpdfstring{\(z_{1:N}^{\mathcal{B}}\)}{z1:N\^{}B})}

\label{app:data_to_particle}

We update each data point's particle assignment \(z_n^{\mathcal{B}}\) for \(n = 1, \dotsc, N\), using the conditional:
\begin{align*}
p(z_n^{\mathcal{B}} = \ell \mid \mathbf{x}_n, \mathbf{v}_n, \text{rest}) &\propto
\pi^{\mathcal{B}}(\ell) \cdot \mathcal{N}(\mathbf{x}_n \mid \boldsymbol{\mu}_\ell^{\mathcal{B}}, \boldsymbol{\Sigma}_\ell^{\mathcal{B}}) \\
&\quad \cdot \mathcal{N}(\mathbf{v}_n \mid \mathbf{v}_\ell, \boldsymbol{\Sigma}_\ell^{\mathcal{V}})
\end{align*}
The prior is given by categorical weights \(\pi^{\mathcal{B}}\); the likelihood is a product of two Gaussians over position \(\mathbf{x}_n\) and velocity \(\mathbf{v}_n\). We compute unnormalized log-probabilities \(\tilde{p}_{n,\ell}\) for each particle:
\begin{align*}
\tilde{p}_{n,\ell} &= \log \pi^{\mathcal{B}}(\ell) + \log \mathcal{N}(\mathbf{x}_n \mid \boldsymbol{\mu}_\ell^{\mathcal{B}}, \boldsymbol{\Sigma}_\ell^{\mathcal{B}}) \\
&\quad + \log \mathcal{N}(\mathbf{v}_n \mid \mathbf{v}_\ell, \boldsymbol{\Sigma}_\ell^{\mathcal{V}})
\end{align*}
and normalize to obtain the categorical:
\[
p(z_n^{\mathcal{B}} = \ell) = \frac{\exp(\tilde{p}_{n,\ell})}{\sum_{\ell'=1}^L \exp(\tilde{p}_{n,\ell'})}
\]
from which we sample:
\[
z_n^{\mathcal{B}} \sim \text{Categorical}(p(z_n^{\mathcal{B}} = 1), \dotsc, p(z_n^{\mathcal{B}} = L))
\]

All data points are jointly reassigned in a blocked manner, each selecting the particle that best explains its position and motion, weighted by the prior over particles.

\subsubsection{Particle Mixture Weights \texorpdfstring{\(\boldsymbol{\pi}^{\mathcal{B}}\)}{pi\^{}B}}

\label{app:particle_weights}

We update the particle mixture weights \(\boldsymbol{\pi}^{\mathcal{B}}\) conditioned on data point-to-particle assignments \(\{z_n^{\mathcal{B}}\}\). By Dirichlet–Categorical conjugacy, the conditional distribution becomes:
\[
\boldsymbol{\pi}^{\mathcal{B}} \mid \{z_n^{\mathcal{B}}\} \sim \mathrm{Dir}(\beta_1 + M_1, \dots, \beta_L + M_L)
\]
where \(M_\ell = \#\{ n : z_n^{\mathcal{B}} = \ell \}\) counts how many data points are currently assigned to each particle \(\ell\). This step re-weights the prior particle proportions according to updated data point assignments.

\subsubsection{Particle Spatial Means \texorpdfstring{\(\boldsymbol{\mu}_\ell^{\mathcal{B}}\)}{mu\_l\^{}B}}

\label{app:particle_spatial_means}

We update each particle center \(\boldsymbol{\mu}_\ell^{\mathcal{B}}\) from its Gaussian conditional, combining:
(1) a spatial prior from its assigned cluster,
(2) position likelihoods from assigned data points, and
(3) a velocity constraint derived from rigid motion.

Let \(\mathbf{A}_\ell = \mathbf{R}_{z_\ell^{\mathcal{H}}} - \mathbf{I}\) and \(\mathbf{b}_\ell = \mathbf{t}_{z_\ell^{\mathcal{H}}} - \mathbf{A}_\ell \boldsymbol{\mu}^{\mathcal{H}}_{z_\ell^{\mathcal{H}}}\). Then:

\[
\mathbf{v}_\ell \sim \mathcal{N}(\mathbf{A}_\ell \boldsymbol{\mu}_\ell^{\mathcal{B}} + \mathbf{b}_\ell, \sigma_V^2 \mathbf{I})
\]
The conditional distribution is a Gaussian-Gaussian conjugate of the form:
\begin{align*}
\boldsymbol{\mu}_\ell^{\mathcal{B}} \mid &\boldsymbol{\mu}^{\mathcal{H}}_{z_\ell^{\mathcal{H}}}, \boldsymbol{\Sigma}^{\mathcal{H}}_{z_\ell^{\mathcal{H}}}, \mathbf{v}_\ell, \mathbf{t}_{z_\ell^{\mathcal{H}}}, \mathbf{R}_{z_\ell^{\mathcal{H}}}, \sigma_V^2, \\
&\{\mathbf{x}_n : z_n^{\mathcal{B}} = \ell\}, \boldsymbol{\Sigma}_\ell^{\mathcal{B}} \sim \mathcal{N}(\mathbf{P}_\ell^{-1} \mathbf{m}_\ell, \mathbf{P}_\ell^{-1})
\end{align*}
with precision and mean:
\begin{align*}
\mathbf{P}_\ell &= (\boldsymbol{\Sigma}^{\mathcal{H}}_{z_\ell^{\mathcal{H}}})^{-1} + N_\ell (\boldsymbol{\Sigma}_\ell^{\mathcal{B}})^{-1} + \frac{1}{\sigma_V^2} \mathbf{A}_\ell^\top \mathbf{A}_\ell \\
\mathbf{m}_\ell &= (\boldsymbol{\Sigma}^{\mathcal{H}}_{z_\ell^{\mathcal{H}}})^{-1} \boldsymbol{\mu}^{\mathcal{H}}_{z_\ell^{\mathcal{H}}} + (\boldsymbol{\Sigma}_\ell^{\mathcal{B}})^{-1} \mathbf{S}_\ell \\
&\quad + \frac{1}{\sigma_V^2} \mathbf{A}_\ell^\top (\mathbf{v}_\ell - \mathbf{b}_\ell),
\end{align*}
where \(N_\ell\) is the number of data points assigned to particle \(\ell\), and \(\mathbf{S}_\ell = \sum_{n: z_n^{\mathcal{B}} = \ell} \mathbf{x}_n\) is the sum of their positions.

\subsubsection{Particle Spatial Covariances \texorpdfstring{\(\boldsymbol{\Sigma}_\ell^{\mathcal{B}}\)}{Sigma\_l\^{}B}}

\label{app:particle_covs}

We update each particle's spatial covariance matrix \(\boldsymbol{\Sigma}_\ell^{\mathcal{B}}\) using Normal–Inverse-Wishart conjugacy. Let \(N_\ell = \#\{n : z_n^{\mathcal{B}} = \ell\}\) be the number of data points assigned to particle \(\ell\), and define the scatter matrix:
\[
\mathbf{S}_\ell = \sum_{n: z_n^{\mathcal{B}} = \ell} (\mathbf{x}_n - \boldsymbol{\mu}_\ell^{\mathcal{B}})(\mathbf{x}_n - \boldsymbol{\mu}_\ell^{\mathcal{B}})^\top
\]
Given an Inverse-Wishart prior \(\mathcal{W}^{-1}(\boldsymbol{\Psi}^{\mathcal{B}}, \nu^{\mathcal{B}})\), the conditional distribution is:
\[
\boldsymbol{\Sigma}_\ell^{\mathcal{B}} \mid \boldsymbol{\mu}_\ell^{\mathcal{B}}, \{\mathbf{x}_n : z_n^{\mathcal{B}} = \ell\} \sim \mathcal{W}^{-1}(\boldsymbol{\Psi}_\ell' = \boldsymbol{\Psi}^{\mathcal{B}} + \mathbf{S}_\ell, \nu^{\mathcal{B}} + N_\ell)
\]
This update adjusts each particle's spatial uncertainty based on the observed spread of its assigned data points.

\subsubsection{Particle Velocity Means \texorpdfstring{\(\mathbf{v}_\ell\)}{v\_l}}

\label{app:particle_vel_means}

We update each particle velocity anchor \(\mathbf{v}_\ell\) via a Gaussian conditional distribution combining: (1) a rigid motion prior from its assigned cluster, and (2) velocity observations from assigned data points. Let \(\bar{\mathbf{v}}_\ell = \mathbf{t}_{z_\ell^{\mathcal{H}}} + (\mathbf{R}_{z_\ell^{\mathcal{H}}} - \mathbf{I}) (\boldsymbol{\mu}_\ell^{\mathcal{B}} - \boldsymbol{\mu}^{\mathcal{H}}_{z_\ell^{\mathcal{H}}})\) be the prior mean.

Given the set \(\{\mathbf{v}_n : z_n^{\mathcal{B}} = \ell\}\) and count \(N_\ell = \#\{n : z_n^{\mathcal{B}} = \ell\}\), the conditional is a Gaussian-Gaussian conjugate update:
\[
\mathbf{v}_\ell \mid \bar{\mathbf{v}}_\ell, \sigma_V^2, \boldsymbol{\Sigma}_\ell^{\mathcal{V}}, \{\mathbf{v}_n : z_n^{\mathcal{B}} = \ell\} \sim \mathcal{N}(\boldsymbol{\mu}_\ell^v, \boldsymbol{\Sigma}_\ell^v)
\]
with:
\begin{align*}
(\boldsymbol{\Sigma}_\ell^v)^{-1} &= \frac{1}{\sigma_V^2} \mathbf{I} + N_\ell (\boldsymbol{\Sigma}_\ell^{\mathcal{V}})^{-1} \\
\boldsymbol{\mu}_\ell^v &= \boldsymbol{\Sigma}_\ell^v \left( \frac{1}{\sigma_V^2} \bar{\mathbf{v}}_\ell + (\boldsymbol{\Sigma}_\ell^{\mathcal{V}})^{-1} \sum_{n:z_n^{\mathcal{B}}=\ell} \mathbf{v}_n \right)
\end{align*}
This update accounts for the velocity prediction from the cluster's rigid transform along with the empirical data point velocities, with each contribution weighted by its respective uncertainty.

\subsubsection{Particle Velocity Covariances \texorpdfstring{\(\boldsymbol{\Sigma}_\ell^{\mathcal{V}}\)}{Sigma\_l\^{}V}}

\label{app:particle_vel_covs}

Each particle's velocity covariance \(\boldsymbol{\Sigma}_\ell^{\mathcal{V}}\) is inferred using Normal–Inverse-Wishart conjugacy. Let \(N_\ell = \#\{n : z_n^{\mathcal{B}} = \ell\}\) be the number of data points assigned to particle \(\ell\), and define the velocity scatter:
\[
\mathbf{T}_\ell = \sum_{n : z_n^{\mathcal{B}} = \ell} (\mathbf{v}_n - \mathbf{v}_\ell)(\mathbf{v}_n - \mathbf{v}_\ell)^\top
\]
Given prior \(\mathcal{W}^{-1}(\boldsymbol{\Psi}^{\mathcal{V}}, \nu^{\mathcal{V}})\), the conditional distribution is:
\[
\boldsymbol{\Sigma}_\ell^{\mathcal{V}} \mid \mathbf{v}_\ell, \{\mathbf{v}_n : z_n^{\mathcal{B}} = \ell\} \sim \mathcal{W}^{-1}(\boldsymbol{\Psi}_\ell' = \boldsymbol{\Psi}^{\mathcal{V}} + \mathbf{T}_\ell,\; \nu^{\mathcal{V}} + N_\ell)
\]
This update reflects the velocity noise structure within each particle, accounting for spread in assigned data point velocities.

\subsubsection{Particle-to-Cluster Assignments (\texorpdfstring{\(z_{1:L}^{\mathcal{H}}\)}{z1:L\^{}H})}

\label{app:particle_to_cluster}

We update each particle's cluster assignment \(z_\ell^{\mathcal{H}}\) for \(\ell = 1, \dotsc, L\), using the conditional:
\begin{align*}
p(z_\ell^{\mathcal{H}} = k \mid \boldsymbol{\mu}_\ell^{\mathcal{B}}, \mathbf{v}_\ell, \text{rest}) &\propto
\pi^{\mathcal{H}}(k) \cdot \mathcal{N}(\boldsymbol{\mu}_\ell^{\mathcal{B}} \mid \boldsymbol{\mu}_k^{\mathcal{H}}, \boldsymbol{\Sigma}_k^{\mathcal{H}}) \\
&\quad \cdot \mathcal{N}\big(\mathbf{v}_\ell \mid \mathbf{t}_k + (\mathbf{R}_k - \mathbf{I}) \\
&\qquad \times (\boldsymbol{\mu}_\ell^{\mathcal{B}} - \boldsymbol{\mu}_k^{\mathcal{H}}), \sigma_V^2 \mathbf{I}\big)
\end{align*}
The prior is given by categorical weights \(\pi^{\mathcal{H}}\); the likelihood combines a spatial Gaussian over the particle's position \(\boldsymbol{\mu}_\ell^{\mathcal{B}}\) and a velocity Gaussian that accounts for rigid-body motion induced by the cluster's rotation \(\mathbf{R}_k\) and translation \(\mathbf{t}_k\). We compute unnormalized log-probabilities \(\tilde{p}_{\ell,k}\) for each cluster:
\begin{align*}
\tilde{p}_{\ell,k} &=
\log \pi^{\mathcal{H}}(k) +
\log \mathcal{N}(\boldsymbol{\mu}_\ell^{\mathcal{B}} \mid \boldsymbol{\mu}_k^{\mathcal{H}}, \boldsymbol{\Sigma}_k^{\mathcal{H}}) \\
&\quad + \log \mathcal{N}\big(\mathbf{v}_\ell \mid \mathbf{t}_k + (\mathbf{R}_k - \mathbf{I})(\boldsymbol{\mu}_\ell^{\mathcal{B}} - \boldsymbol{\mu}_k^{\mathcal{H}}), \sigma_V^2 \mathbf{I}\big)
\end{align*}
and normalize to obtain the categorical:
\[
p(z_\ell^{\mathcal{H}} = k) = \frac{\exp(\tilde{p}_{\ell,k})}{\sum_{k'=1}^K \exp(\tilde{p}_{\ell,k'})}
\]
from which we sample:
\[
z_\ell^{\mathcal{H}} \sim \text{Categorical}(p(z_\ell^{\mathcal{H}} = 1), \dotsc, p(z_\ell^{\mathcal{H}} = K))
\]
This constitutes a blocked Gibbs step, where all particle-to-cluster assignments are jointly updated. Each particle selects the cluster whose spatial and rigid motion parameters best explain its position and velocity.

\subsubsection{Cluster Mixture Weights \texorpdfstring{\(\boldsymbol{\pi}^{\mathcal{H}}\)}{pi\^{}H}}

\label{app:cluster_weights}

We update the cluster mixture weights \(\boldsymbol{\pi}^{\mathcal{H}}\) given particle-to-cluster assignments \(\{z_\ell^{\mathcal{H}}\}\). Using Dirichlet–Categorical conjugacy, the conditional is:
\[
\boldsymbol{\pi}^{\mathcal{H}} \mid \{z_\ell^{\mathcal{H}}\} \sim \mathrm{Dir}(\alpha_1 + N_1, \dots, \alpha_K + N_K)
\]
where \(N_k = \#\{ \ell : z_\ell^{\mathcal{H}} = k \}\) is the number of particles assigned to cluster \(k\). This step updates the prior cluster proportions based on current assignment counts.

\subsubsection{Cluster Spatial Means \texorpdfstring{\(\boldsymbol{\mu}_k^{\mathcal{H}}\)}{mu\_k\^{}H}}

\label{app:cluster_means}

We update each cluster center \(\boldsymbol{\mu}_k^{\mathcal{H}}\) via a Gaussian conditional that integrates: (1) a Gaussian prior centered at \(\mu^{\mathcal{H}}\), (2) assigned particle centers \(\boldsymbol{\mu}_\ell^{\mathcal{B}}\), and (3) observed particle velocities corrected by the cluster's affine transform.

Let \(\mathbf{A}_k = \mathbf{I} - \mathbf{R}_k\) and \(\mathbf{b}_\ell = \mathbf{t}_k - \mathbf{A}_k \boldsymbol{\mu}_\ell^{\mathcal{B}}\). Then the velocity residual is:
\[
\mathbf{r}_\ell = \mathbf{v}_\ell - \mathbf{b}_\ell
\]
Given the sum of assigned particle means \(\mathbf{S}_k = \sum_{\ell : z_\ell^{\mathcal{H}} = k} \boldsymbol{\mu}_\ell^{\mathcal{B}}\), the velocity residual sum \(\mathbf{R}_k = \sum_{\ell : z_\ell^{\mathcal{H}} = k} \mathbf{r}_\ell\), and the count \(N_k = \#\{\ell : z_\ell^{\mathcal{H}} = k\}\) of particles assigned to cluster \(k\), the conditional is:

\begin{align*}
\boldsymbol{\mu}_k^{\mathcal{H}} \mid & \mu^{\mathcal{H}}, \sigma_H^2, \boldsymbol{\Sigma}_k^{\mathcal{H}}, \mathbf{t}_k, \mathbf{R}_k, \sigma_V^2, \mathbf{R}_k, \\
&\{\boldsymbol{\mu}_\ell^{\mathcal{B}}, \mathbf{v}_\ell : z_\ell^{\mathcal{H}} = k\}
\sim \mathcal{N}(P_k^{-1} \mathbf{m}_k, P_k^{-1})
\end{align*}

with:
\begin{align*}
P_k &= \frac{1}{\sigma_H^2} \mathbf{I} + N_k \left( \boldsymbol{\Sigma}_k^{\mathcal{H}\,-1} + \frac{1}{\sigma_V^2} \mathbf{A}_k^\top \mathbf{A}_k \right) \\
\mathbf{m}_k &= \frac{1}{\sigma_H^2} \mu^{\mathcal{H}} + \boldsymbol{\Sigma}_k^{\mathcal{H}\,-1} \mathbf{S}_k + \frac{1}{\sigma_V^2} \mathbf{A}_k^\top \mathbf{R}_k
\end{align*}
This update integrates global priors, spatial evidence from assigned particles, and velocity-based constraints under rigid motion. We parallelize this step by batching cluster-level quantities over \(K\) and particle-level inputs over \(L\), with per-cluster residual aggregation. The final blocked multivariate normal update samples new cluster means in parallel from their respective posteriors.

\subsubsection{Cluster Spatial Covariances \texorpdfstring{\(\boldsymbol{\Sigma}_k^{\mathcal{H}}\)}{Sigma\_k\^{}H}}

\label{app:cluster_covs}

We infer each cluster's spatial covariance \(\boldsymbol{\Sigma}_k^{\mathcal{H}}\) using a Normal–Inverse-Wishart update conditioned on its assigned particles. Let \(L_k = \#\{\ell : z_\ell^{\mathcal{H}} = k\}\) be the number of particles assigned to cluster \(k\), and define the cluster-centered scatter:
\[
\mathbf{S}_k = \sum_{\ell : z_\ell^{\mathcal{H}} = k} (\boldsymbol{\mu}_\ell^{\mathcal{B}} - \boldsymbol{\mu}_k^{\mathcal{H}})(\boldsymbol{\mu}_\ell^{\mathcal{B}} - \boldsymbol{\mu}_k^{\mathcal{H}})^\top
\]
Given the Inverse-Wishart prior \(\mathcal{W}^{-1}(\boldsymbol{\Psi}^{\mathcal{H}}, \nu^{\mathcal{H}})\), the conditional becomes, with \(\boldsymbol{\Psi}_k' = \boldsymbol{\Psi}^{\mathcal{H}} + \mathbf{S}_k\) and \(\nu_k' = \nu^{\mathcal{H}} + L_k\):
\[
\boldsymbol{\Sigma}_k^{\mathcal{H}} \mid \boldsymbol{\mu}_k^{\mathcal{H}}, \{\boldsymbol{\mu}_\ell^{\mathcal{B}} : z_\ell^{\mathcal{H}} = k\} \sim \mathcal{W}^{-1}(\boldsymbol{\Psi}_k', \nu_k')
\]
This posterior captures the spatial extent of each cluster based on the spread of its assigned particle centers.

\subsubsection{Cluster Rotation \texorpdfstring{\(\mathbf{R}_k\)}{R\_k}}

\label{app:cluster_rot}

We update each cluster's rotation matrix \(\mathbf{R}_k\) by evaluating a discrete set of candidate rotations \(\{ \mathbf{R}^{(j)} \}_{j=1}^{M_r}\) drawn from a spherical cap (e.g., von Mises–Fisher). For each candidate, we compute a probability based on how well the induced rigid motion explains observed particle velocities. Let \(\bar{\mathbf{v}}_\ell^{(j)} = \mathbf{t}_k + (\mathbf{R}^{(j)} - \mathbf{I})(\boldsymbol{\mu}_\ell^{\mathcal{B}} - \boldsymbol{\mu}_k^{\mathcal{H}})\) be the expected velocity for particle \(\ell\) under candidate \(j\). Then:
\[
\log \tilde{q}_j = \sum_{\ell : z_\ell^{\mathcal{H}} = k} \log \mathcal{N}(\mathbf{v}_\ell \mid \bar{\mathbf{v}}_\ell^{(j)}, \sigma_V^2 \mathbf{I})
\]
Adding the prior log-probabilities \(\log p(\mathbf{R}^{(j)})\), we normalize the log-scores to obtain:
\[
q_j = \frac{\exp(\log \tilde{q}_j + \log p(\mathbf{R}^{(j)}))}{\sum_{j'=1}^{M_r} \exp(\log \tilde{q}_{j'} + \log p(\mathbf{R}^{(j')}))}
\]
from which we sample:
\[
\mathbf{R}_k \sim \text{Categorical}(\{q_j\}_{j=1}^{M_r})
\]
This update selects the rotation that best aligns relative particle positions with their observed velocities, conditioned on the current cluster translation \(\mathbf{t}_k\), velocity noise $\sigma_V^2$, cluster means ($\boldsymbol{\mu}_k^{\mathcal{H}}$) and assigned particle means ($\{\boldsymbol{\mu}_\ell^{\mathcal{B}} : z_\ell^{\mathcal{H}} = k\}$).

\subsubsection{Cluster Translation Velocities \texorpdfstring{\(\mathbf{t}_k\)}{t\_k}}

\label{app:cluster_trans}

We update each cluster's translation velocity \(\mathbf{t}_k\) by evaluating a discrete set of candidate translations \(\{\mathbf{t}^{(m)}\}_{m=1}^{M_t}\) sampled from an isotropic Gaussian prior \(\mathcal{N}(\mathbf{0}, s^2 \mathbf{I})\). Each candidate is scored based on how well it explains the observed particle velocities under the current rotation \(\mathbf{R}_k\). Let \(\bar{\mathbf{v}}_\ell^{(m)} = \mathbf{t}^{(m)} + (\mathbf{R}_k - \mathbf{I})(\boldsymbol{\mu}_\ell^{\mathcal{B}} - \boldsymbol{\mu}_k^{\mathcal{H}})\) be the expected velocity for particle \(\ell\) under candidate \(m\). Then:
\[
\log \tilde{p}_m = \sum_{\ell : z_\ell^{\mathcal{H}} = k} \log \mathcal{N}(\mathbf{v}_\ell \mid \bar{\mathbf{v}}_\ell^{(m)}, \sigma_V^2 \mathbf{I})
\]
We add prior log-probabilities and normalize to form a categorical:
\[
p_m = \frac{\exp(\log \tilde{p}_m + \log p(\mathbf{t}^{(m)}))}{\sum_{m'=1}^{M_t} \exp(\log \tilde{p}_{m'} + \log p(\mathbf{t}^{(m')}))}
\]
from which we sample:
\[
\mathbf{t}_k \sim \text{Categorical}(\{p_m\}_{m=1}^{M_t})
\]
This update selects the translation that best explains the observed particle velocities, conditioned on current cluster rotation \(\mathbf{R}_k\), velocity noise \(\sigma_V^2\), cluster center \(\boldsymbol{\mu}_k^{\mathcal{H}}\), and assigned particle means \(\{\boldsymbol{\mu}_\ell^{\mathcal{B}} : z_\ell^{\mathcal{H}} = k\}\).

\subsection{Initialization Procedure}
\label{app:gibbs_init}

It is well known that MCMC chains are sensitive to the initialization and should be initialized at a high density region. In both the 2D and 3D variants of GenMatter, we use K-Means clustering and a data-driven approach to initialize the MCMC chain for the initial frame ($T=0$).

\subsubsection{K-Means and Data-driven Initialization at T=0}

\label{app:init_steps}

Given the number of particles ($L$), we use K-means via a K-Means++ initialization to initialize the particle spatial positions ($\boldsymbol{\mu}_\ell^{\mathcal{B}}$). We then use an additional K-means step to initialize the cluster spatial positions ($\boldsymbol{\mu}_k^{\mathcal{H}}$) by treating the particle spatial positions as data points to cluster.

This K-means initialization provides initial values for assignments at both layers (\(z_n^\mathcal{B}\), \(z_\ell^\mathcal{H}\)). We then use these assignments to initialize the mixture weights at both layers (\(\pi^{\mathcal{B}}\), \(\pi^{\mathcal{H}}\)) by computing the empirical frequencies of each cluster and normalizing: \(\pi^{\mathcal{B}}_\ell = \frac{M_\ell}{N}\) and \(\pi^{\mathcal{H}}_k = \frac{N_k}{L}\), where \(M_\ell\) is the number of datapoints assigned to particle \(\ell\) and \(N_k\) is the number of particles assigned to cluster \(k\). We initialize the velocity mean of each particle \(\mathbf{v}_\ell\) by averaging the observed velocities of the datapoints assigned to it:
\[
\mathbf{v}_\ell = \frac{1}{M_\ell} \sum_{n: z_n^{\mathcal{B}} = \ell} \mathbf{v}_n.
\]

To initialize the covariance matrices, we compute the sample covariance of the relevant residuals for each component:

\begin{enumerate}
    \item \textbf{Particle Spatial Covariance:}
    \[
    \boldsymbol{\Sigma}_\ell^{\mathcal{B}} = \frac{1}{M_\ell - 1} \sum_{n: z_n^{\mathcal{B}} = \ell} (\mathbf{x}_n - \boldsymbol{\mu}_\ell^{\mathcal{B}})(\mathbf{x}_n - \boldsymbol{\mu}_\ell^{\mathcal{B}})^\top.
    \]

    \item \textbf{Particle Velocity Covariance:}
    \[
    \boldsymbol{\Sigma}_\ell^{\mathcal{V}} = \frac{1}{M_\ell - 1} \sum_{n: z_n^{\mathcal{B}} = \ell} (\mathbf{v}_n - \mathbf{v}_\ell)(\mathbf{v}_n - \mathbf{v}_\ell)^\top.
    \]

    \item \textbf{Cluster Spatial Covariance:}
    \[
    \boldsymbol{\Sigma}_k^{\mathcal{H}} = \frac{1}{N_k - 1} \sum_{\ell: z_\ell^{\mathcal{H}} = k} (\boldsymbol{\mu}_\ell^{\mathcal{B}} - \boldsymbol{\mu}_k^{\mathcal{H}})(\boldsymbol{\mu}_\ell^{\mathcal{B}} - \boldsymbol{\mu}_k^{\mathcal{H}})^\top.
    \]
\end{enumerate}

To initialize each cluster's rigid transform \((\mathbf{R}_k, \mathbf{t}_k)\), we apply the Kabsch algorithm to align assigned particle positions with their next-frame displacements. For cluster \(k\), we collect all datapoints \(\mathbf{x}_n\) assigned to particles \(\ell\) with \(z_\ell^{\mathcal{H}} = k\) and define their estimated displacements \(\mathbf{x}_n' = \mathbf{x}_n + \mathbf{v}_n\). Let \(\mathcal{X}_k = \{\mathbf{x}_n\}\) and \(\mathcal{X}_k' = \{\mathbf{x}_n'\}\) be the source and target sets.

We compute centroids \(\bar{\mathbf{x}}_k = \frac{1}{|\mathcal{X}_k|} \sum \mathbf{x}_n\), \(\bar{\mathbf{x}}_k' = \frac{1}{|\mathcal{X}_k'|} \sum \mathbf{x}_n'\), and form centered sets \(\tilde{\mathbf{x}}_n = \mathbf{x}_n - \bar{\mathbf{x}}_k\), \(\tilde{\mathbf{x}}_n' = \mathbf{x}_n' - \bar{\mathbf{x}}_k'\). The cross-covariance matrix is:
\[
\mathbf{H}_k = \sum_n \tilde{\mathbf{x}}_n \tilde{\mathbf{x}}_n'^\top
\]

We compute the singular value decomposition \(\mathbf{H}_k = \mathbf{U}_k \boldsymbol{\Sigma}_k \mathbf{V}_k^\top\), and define the optimal rotation as:
\[
\mathbf{R}_k = \mathbf{V}_k \mathbf{D}_k \mathbf{U}_k^\top
\]
where \(\mathbf{D}_k\) is  defined as:
\[
\mathbf{D}_k =
\begin{cases}
\begin{bmatrix}
1 & 0 \\
0 & \det(\mathbf{V}_k \mathbf{U}_k^\top)
\end{bmatrix} & \text{(2D model)} \\[12pt]
\begin{bmatrix}
1 & 0 & 0 \\
0 & 1 & 0 \\
0 & 0 & \det(\mathbf{V}_k \mathbf{U}_k^\top)
\end{bmatrix} & \text{(3D model)}
\end{cases}
\]

The corresponding translation is:
\[
\mathbf{t}_k = \bar{\mathbf{x}}_k' - \mathbf{R}_k \bar{\mathbf{x}}_k
\]

This provides an initialization of cluster motion consistent with the observed displacements of assigned particles. The update is applied independently for each cluster \(k = 1, \dotsc, K\).

\subsubsection{Data-Dependent Hyperparameters}

We initialize model hyperparameters directly from empirical statistics computed on the initial frame ($T = 0$). The global cluster location prior \(\mu^{\mathcal{H}}\) is set to the median datapoint position, while the prior spatial scale \(\boldsymbol{\Psi}^{\mathcal{B}}, \boldsymbol{\Psi}^{\mathcal{H}}, \boldsymbol{\Psi}^{\mathcal{V}}\) are initialized using the median initialized particle and cluster covariances length scales.

The degrees of freedom \(\nu^{\mathcal{B}}, \nu^{\mathcal{H}}, \nu^{\mathcal{V}}\) are initialized proportionally to the number of datapoints assigned, weighted by particle or cluster weights:
\begin{align*}
\nu^{\mathcal{B}} &= \left\lfloor \mathrm{median}(w_\ell^{\mathcal{B}} \cdot N) \right\rfloor, \\
\nu^{\mathcal{H}} &= \left\lfloor \mathrm{median}(w_k^{\mathcal{H}} \cdot N) \right\rfloor, \\
\nu^{\mathcal{V}} &= \left\lfloor \mathrm{median}(w_\ell^{\mathcal{B}} \cdot N) \right\rfloor
\end{align*}
where \(w_\ell^{\mathcal{B}}\) and \(w_k^{\mathcal{H}}\) are the normalized empirical weights of each particle and cluster.



\subsection{Tracking Gibbs Procedure}
\label{app:gibbs_tracking}

To perform inference over video sequences, we extend our generative particle model into the sequential filtering regime using a structured Markov Chain Monte Carlo (MCMC) procedure. Specifically, we implement a blocked Gibbs sampler that leverages the causal ordering of the variables from the previous frame to initialize each frame and performs bottom-up inference to refine all data point-, particle-, and cluster-level variables. Our approach maintains a tractable posterior approximation at each timestep by propagating forward a subset of latent variables and resampling the remaining ones conditioned on new observations. This sequential per-frame MCMC design supports inference in dynamic scenes where data associations must be re-inferred at every timestep.

At each timestep $t$, we target the posterior over latent variables given the observed data point positions $\mathbf{x}_{1:N}^t$ and velocities $\mathbf{v}_{1:N}^t$:
\begin{align*}
p(&\boldsymbol{\mu}_\mathcal{H}^t, \boldsymbol{\Sigma}_\mathcal{H}^t, \mathbf{R}_\mathcal{H}^t, \mathbf{t}_\mathcal{H}^t,
\boldsymbol{\mu}_\mathcal{B}^t, \mathbf{v}_\mathcal{B}^t, \boldsymbol{\Sigma}_\mathcal{V}^t, \\
&z_{1:N}^t, z_{1:L}^t, \boldsymbol{\pi}_\mathcal{B}^t, \boldsymbol{\pi}_\mathcal{H}^t \mid \mathbf{x}_{1:N}^t, \mathbf{v}_{1:N}^t)
\end{align*}
where $\boldsymbol{\Sigma}_\mathcal{B}$ (particle spatial covariances) are held fixed throughout tracking in both 2D and 3D experiments to preserve the spatial extent of the deformable visual matter represented by each particle, and particle-to-cluster assignments $z_{1:L}^t$ are held fixed only in the 3D case to keep consistency with the initial scene segmentation.

\paragraph{Particle Propagation and Initialization}
Each frame begins by propagating the inferred particle means using their previously inferred velocity vectors:

\begin{equation*}
\tilde{\boldsymbol{\mu}}_\ell^{\mathcal{B}, t} = \boldsymbol{\mu}_\ell^{\mathcal{B}, t{-}1} + \mathbf{v}_\ell^{t{-}1}
\end{equation*}
This serves as an initialization for the particle positions in the next frame.

\paragraph{First Assignment: Spatial Anchoring}
Data points are first assigned to particles based on spatial likelihoods alone:
\begin{equation*}
p(z_n^{\mathcal{B}, t} = \ell \mid \mathbf{x}_n^t) \propto
\pi_\ell^{\mathcal{B}} \cdot \mathcal{N}(\mathbf{x}_n^t \mid \tilde{\boldsymbol{\mu}}_\ell^{\mathcal{B}, t}, \boldsymbol{\Sigma}_\ell^{\mathcal{B}})
\end{equation*}
This step is crucial because, in the absence of known correspondences across frames, we cannot assume that data point $n$ at time $t{-}1$ is the same as datapoint $n$ at time $t$. Instead, we reinterpret each new frame as an unordered set of observations and rely on spatial proximity to propagated particle means to re-establish associations. By using position alone and excluding any top-down beliefs from velocity or cluster structure, this step provides a stable initialization for the rest of the Gibbs updates. Note that this is a partial version of the full assignment step described in Appendix~\ref{app:data_to_particle}, used here to anchor the initial framewise alignment. After assignments, we update the mixture weights $\boldsymbol{\pi}^{\mathcal{B}}$ by sampling from their conjugate Dirichlet distribution (Appendix~\ref{app:particle_weights}).

\paragraph{Particle Mean Update}
After data points have been assigned to particles based on spatial proximity, we update each particle's spatial mean to better reflect this assignment. Specifically, we sample the particle mean from its posterior conditioned on the assigned data points and the expected motion induced by its cluster assignment, as detailed in Appendix~\ref{app:particle_spatial_means}. Since the assignments in the previous Gibbs step compensate for the absence of point-wise correspondences, this update typically results in small adjustments to the propagated means, ensuring that particles remain anchored to observed data while maintaining temporal coherence with the previous frame.

\paragraph{Second Assignment and Particle Refinement}
A second data point-to-particle assignment uses both spatial and velocity likelihoods as described in Appendix~\ref{app:data_to_particle}:
\begin{align*}
p(z_n^{\mathcal{B}, t} = \ell \mid \mathbf{x}_n^t, \mathbf{v}_n^t) &\propto
\pi_\ell^{\mathcal{B}} \cdot
\mathcal{N}(\mathbf{x}_n^t \mid \boldsymbol{\mu}_\ell^{\mathcal{B}}, \boldsymbol{\Sigma}_\ell^{\mathcal{B}}) \\
&\quad \cdot \mathcal{N}(\mathbf{v}_n^t \mid \mathbf{v}_\ell, \boldsymbol{\Sigma}_\ell^{\mathcal{V}})
\end{align*}
This step helps resolve ambiguous associations by combining spatial proximity with motion information. The mixture weights $\boldsymbol{\pi}^{\mathcal{B}}$ are updated again based on the refined assignments (Appendix~\ref{app:particle_weights}).

Each particle's velocity mean $\mathbf{v}_\ell$ is updated from its posterior as described in Appendix~\ref{app:particle_vel_means}, and the velocity covariance $\boldsymbol{\Sigma}_\ell^{\mathcal{V}}$ is resampled as shown in Appendix~\ref{app:particle_vel_covs}. These updates reflect the motion structure inferred from grouped data point velocities.

\paragraph{Cluster-level Updates}
Each particle is assigned to a cluster using a joint spatial and velocity likelihood as described in Appendix~\ref{app:particle_to_cluster}, and the cluster mixture weights $\boldsymbol{\pi}^{\mathcal{H}}$ are resampled using the equation in Appendix~\ref{app:cluster_weights}. Conditioned on these assignments, the cluster mean $\boldsymbol{\mu}_k^{\mathcal{H}}$ and spatial covariance $\boldsymbol{\Sigma}_k^{\mathcal{H}}$ are updated from their conditional distributions (Appendix~\ref{app:cluster_means} and~\ref{app:cluster_covs}), and the rigid transform $(\mathbf{R}_k, \mathbf{t}_k)$ is inferred by categorical sampling over candidate rotations and translations (Appendix~\ref{app:cluster_rot} and~\ref{app:cluster_trans}).

In the 3D experiment, particle-to-cluster assignments $z_{1:L}^t$ are held fixed throughout tracking to stabilize the scene representation's semantic content, which provides a reliable prior over object structure. However, cluster parameters including spatial statistics and rigid transforms are still inferred at each frame to update the spatial localization of the structure given in the original segmentation.




\section{Feature-augmented Variant}
\label{app:feature_model}

\subsection{Model Modification and Initialization}

In the feature-augmented variant of our model, we incorporate image features as additional dimensions of the data points. Following the main text, we define augmented data points $\tilde{\mathbf{x}}_n = [\mathbf{x}_n; \mathbf{f}_n]$ where $\mathbf{f}_n$ are feature vectors extracted from the image. We use the first 10 PCA components of DINO features, where the PCA basis is computed by analyzing all per-pixel features across the entire video. Each particle $\ell$ is associated with a feature mean $\mathbf{f}_\ell$, and the sampling process of the data point features $\mathbf{f}_n$ from the particle features is defined as a Gaussian with variance $\sigma_F^2$:

\[
\mathbf{f}_n \sim \mathcal{N}(\mathbf{f}_\ell, \sigma_F^2 \mathbf{I})
\]

We only fit our per-particle feature parameter during initialization. We perform the steps described in Appendix~\ref{app:init_steps}, followed by computing the initial feature mean of each particle $\mathbf{f}_\ell$ as the average feature vector of its assigned data points:

\[
\mathbf{f}_\ell = \frac{1}{M_\ell} \sum_{n : z_n^{\mathcal{B}} = \ell} \mathbf{f}_n
\]
where \(M_\ell = \#\{n : z_n^{\mathcal{B}} = \ell\}\) is the number of data points assigned to particle \(\ell\). This feature mean serves as the representative feature vector for each particle throughout inference.

\subsection{Data point-to-Particle Assignments with Feature Likelihood}

The main modification to the Gibbs sampler involves the data point-to-particle assignment step, which is modified to include feature similarity. We update each data point's particle assignment \(z_n^{\mathcal{B}}\) for \(n = 1, \dotsc, N\), using the conditional distribution:
\begin{align*}
p(z_n^{\mathcal{B}} = \ell \mid \mathbf{x}_n, \mathbf{v}_n, \mathbf{f}_n, \text{rest}) &\propto
\pi^{\mathcal{B}}(\ell) \cdot \mathcal{N}(\mathbf{x}_n \mid \boldsymbol{\mu}_\ell^{\mathcal{B}}, \boldsymbol{\Sigma}_\ell^{\mathcal{B}}) \\
&\quad \cdot \mathcal{N}(\mathbf{v}_n \mid \mathbf{v}_\ell, \boldsymbol{\Sigma}_\ell^{\mathcal{V}}) \\
&\quad \cdot \mathcal{N}(\mathbf{f}_n \mid \mathbf{f}_\ell, \sigma_F^2 \mathbf{I})
\end{align*}

The prior is given by categorical weights \(\pi^{\mathcal{B}}\), and the likelihood now consists of three independent Gaussian terms: one for position \(\mathbf{x}_n\), one for velocity \(\mathbf{v}_n\), and one for features \(\mathbf{f}_n\). The feature likelihood uses an isotropic covariance \(\sigma_F^2 \mathbf{I}\), which assumes that the features are independent and identically distributed.

We compute unnormalized log-probabilities \(\tilde{p}_{n,\ell}\) for each particle:
\begin{align*}
\tilde{p}_{n,\ell} &= \log \pi^{\mathcal{B}}(\ell) +
\log \mathcal{N}(\mathbf{x}_n \mid \boldsymbol{\mu}_\ell^{\mathcal{B}}, \boldsymbol{\Sigma}_\ell^{\mathcal{B}}) \\
&\quad +
\log \mathcal{N}(\mathbf{v}_n \mid \mathbf{v}_\ell, \boldsymbol{\Sigma}_\ell^{\mathcal{V}}) +
\log \mathcal{N}(\mathbf{f}_n \mid \mathbf{f}_\ell, \sigma_F^2 \mathbf{I})
\end{align*}
and normalize to obtain the categorical conditional distribution:
\[
p(z_n^{\mathcal{B}} = \ell) = \frac{\exp(\tilde{p}_{n,\ell})}{\sum_{\ell'=1}^L \exp(\tilde{p}_{n,\ell'})}
\]
from which we sample:
\[
z_n^{\mathcal{B}} \sim \text{Categorical}(p(z_n^{\mathcal{B}} = 1), \dotsc, p(z_n^{\mathcal{B}} = L))
\]
This update is also a blocked update, executed in a computational manner similar to Appendix~\ref{app:data_to_particle}.

\begin{figure*}[h]
    \centering
    \includegraphics[width=0.75\textwidth]{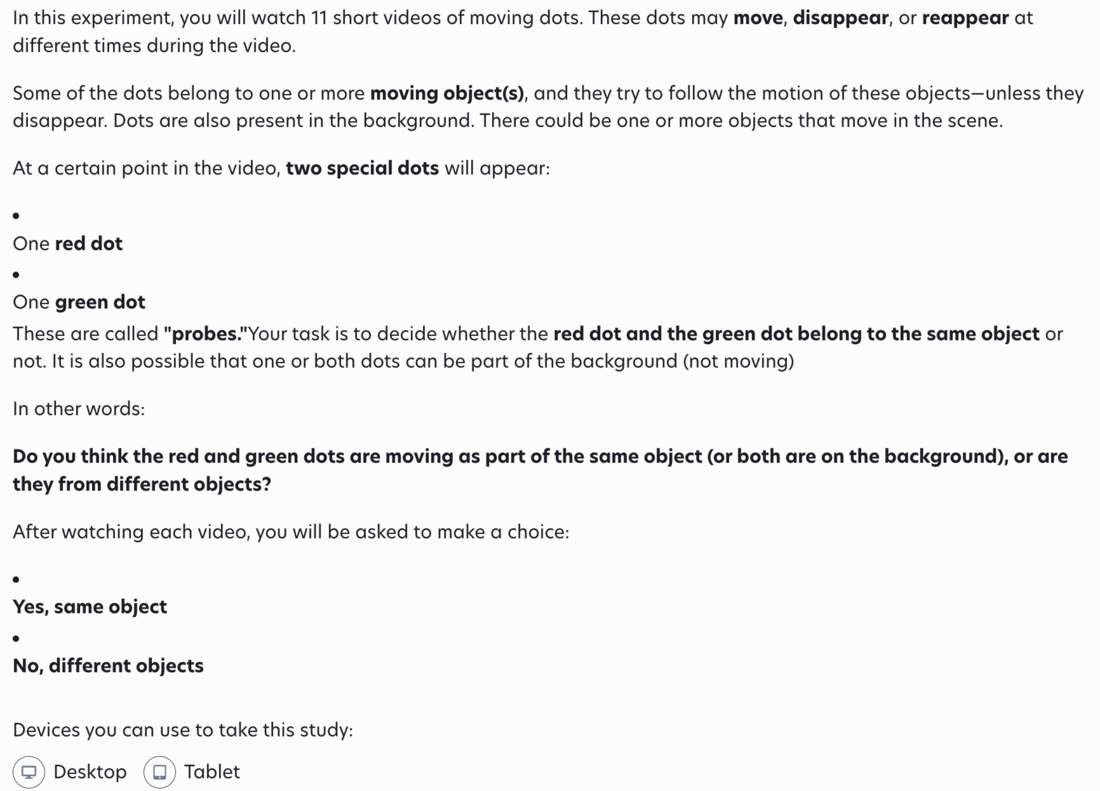}
    \caption{Instructions shown to all participants. This task was allowed to be conducted on either a desktop or a tablet. The 11 videos mentioned refer to the 2 familiarization trials and 9 test trials. Details of compensation is cropped out to preserve anonymity.}
    \label{fig:prolific_inst}
\end{figure*}

\section{Human Psychophysics Experiment}
\label{app:human}

A total of 9 RDKs were created, with each RDK having 3 separate time points and locations where we introduce the red and green dot probes to create a total of 27 stimuli. We recruited a total of 150 human participants through the Prolific platform and all participants were paid at least the local minimum wage for an expected completion time of 4 minutes. The study was designed and conducted under an approved institutional review board (IRB) protocol. All demographic data collected were fully anonymized, and no personally identifying information was provided or collected. All participants were filtered for the following conditions:

\begin{enumerate}
    \item Fluent in English as the study is conducted in English.
    \item Explicitly declared to not have color-blindness, as this study requires each participant to distinguish the red and green probes clearly from the rest of the points in the stimuli.
    \item Has normal to corrected vision, as this study requires clear vision of the stimuli.
\end{enumerate}

The instructions as viewed on Prolific for this study can be seen in Figure~\ref{fig:prolific_inst}. We used Google Forms to conduct the data collection.

\begin{figure*}[h!]
    \centering
    \begin{subfigure}[t]{0.45\textwidth}
    \includegraphics[width=\linewidth]{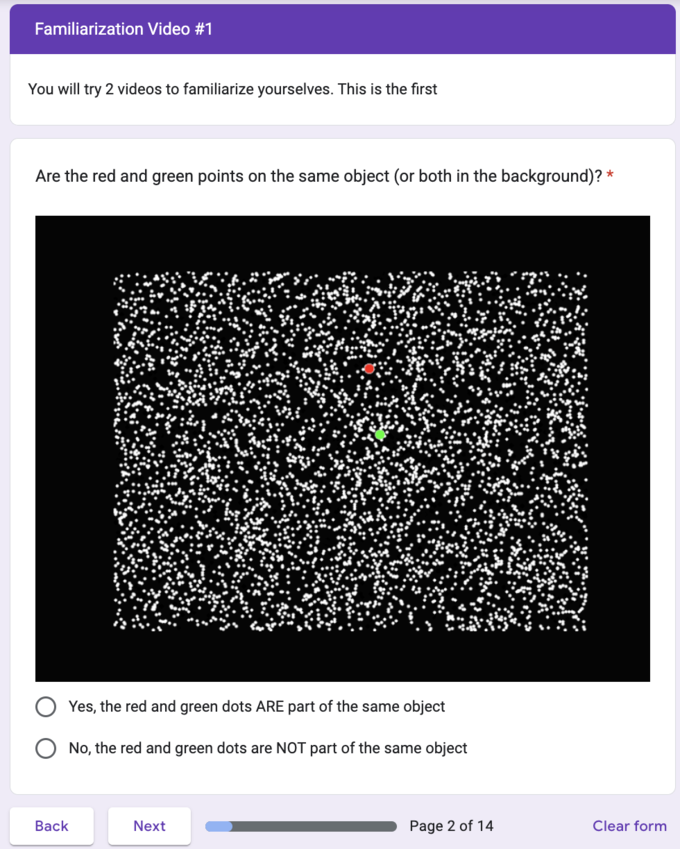}
        \caption{First familiarization trial}
        \label{fig:fam1}
    \end{subfigure}
    \hfill
    \begin{subfigure}[t]{0.45\textwidth}
    \includegraphics[width=\linewidth]{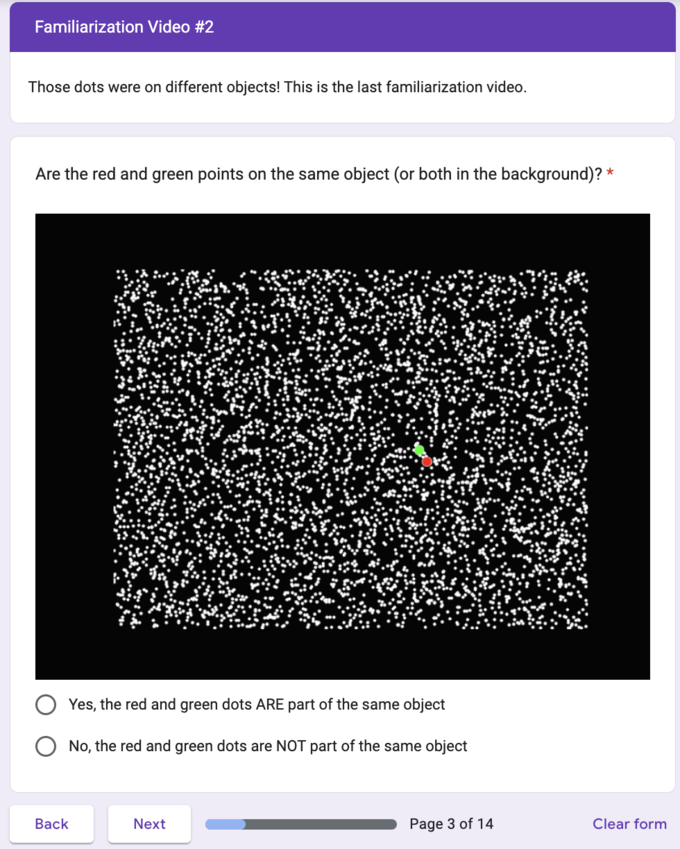}
        \caption{Second familiarization trial with ground truth answer for first familiarization trial revealed.}
        \label{fig:fam2}
    \end{subfigure}
    \hfill
    \begin{subfigure}[t]{0.75\textwidth}
    \includegraphics[width=\linewidth]{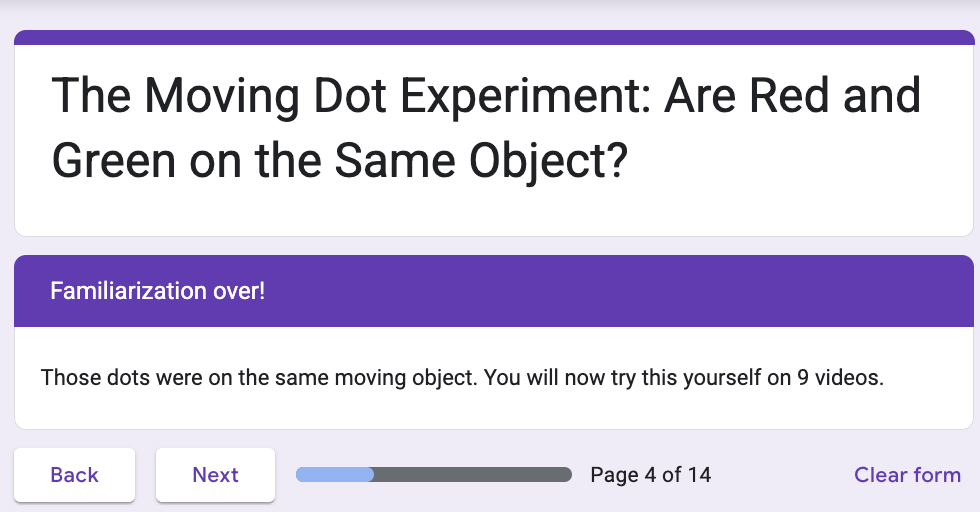}
        \caption{Ground Truth answer for Second Familiarization trial revealed.}
        \label{fig:fam3}
    \end{subfigure}
    \caption{Visual descriptions of the familiarization trials, shown to all 150 participants}
    \label{fig:fam}
\end{figure*}

The instructions were repeated in the Google Form and each participant saw two familiarization trials with feedback on the correct answer. Figure~\ref{fig:fam} shows how these familiarization trials looked to the participant. 


\section{Gestalt 3D Inference}
\label{app:gestalt}

We provide additional implementation details for the Gestalt 3D structure-from-motion experiments described in the main paper.

\noindent
\textbf{Data preprocessing} We compute RAFT optical flow and VideoDepthAnything monocular depth on the native $1000 \times 1000$ frames, then downsample to $96 \times 96$ for inference. We lift 2D pixels to 3D using a pinhole camera model with focal length scaled by $2.0$ to enhance depth separation, and compute 3D motion vectors from optical flow.

\noindent
\textbf{Initialization} GenMatter uses $L = 100$ particles and $K = 5$ clusters. We use the initialization procedure in Appendix~\ref{app:init_steps}, with the addition of a coarse segmentation mask proposal containing all data points with flow magnitude above the median flow magnitude, as well as data points within a standard deviation of the median depth. These assumptions are loose and apply to any natural image regime where GenMatter could be run, and are used only to accelerate MCMC burn-in (since a proposal does not change the posterior being approximated). We run 50 Gibbs sweeps on frame 0 to initialize all model parameters.

\noindent
\textbf{Per-frame Gibbs schedule} For tracking frames $t = 1, \ldots, 4$, we apply 20 Gibbs iterations focused on velocity parameters, followed by 500 full Gibbs sweeps. Particle-to-cluster assignments and particle spatial covariances remain fixed throughout tracking, but data point-to-particle assignments are resampled at each frame.

\begin{figure*}[h!]
    \centering
    \setlength{\tabcolsep}{8pt}
    \renewcommand{\arraystretch}{0}
    \begin{tabular}{@{}ccc@{}}
    \includegraphics[width=0.3\textwidth]{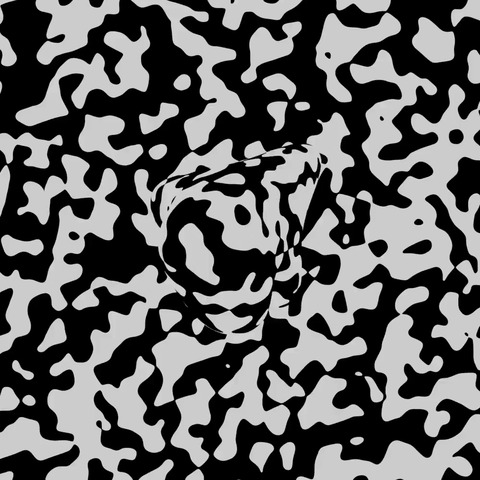} &
    \includegraphics[width=0.3\textwidth]{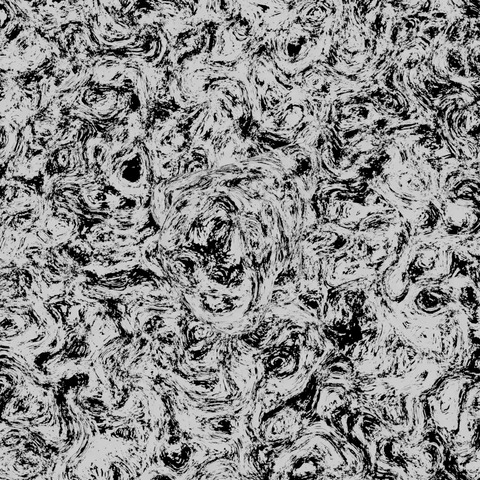} &
    \includegraphics[width=0.3\textwidth]{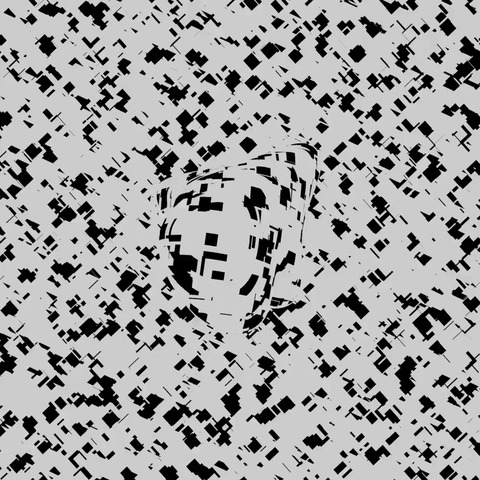} \\[0.15cm]
    \normalsize Texture 00 & \normalsize Texture 07 & \normalsize Texture 13 \\[0.3cm]
    \end{tabular}
    \begin{tabular}{@{}cc@{}}
    \includegraphics[width=0.3\textwidth]{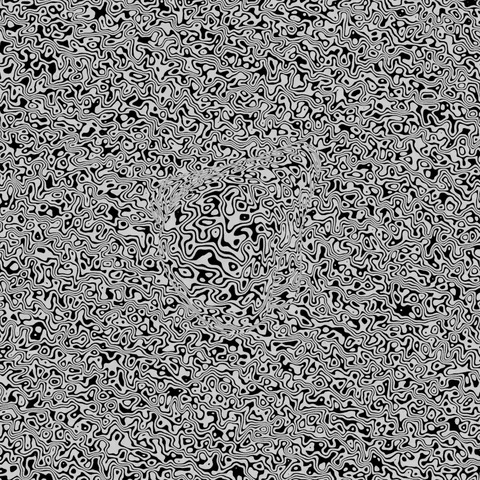} &
    \includegraphics[width=0.3\textwidth]{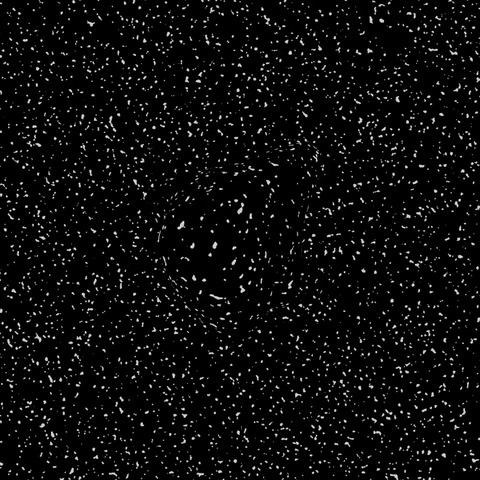} \\[0.15cm]
    \normalsize Texture 16 & \normalsize Texture 21 \\[0.3cm]
    \end{tabular}
    \begin{tabular}{@{}cc@{}}
    \includegraphics[width=0.3\textwidth]{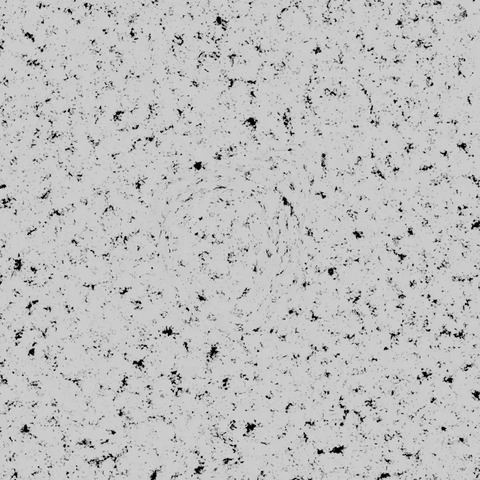} &
    \includegraphics[width=0.3\textwidth]{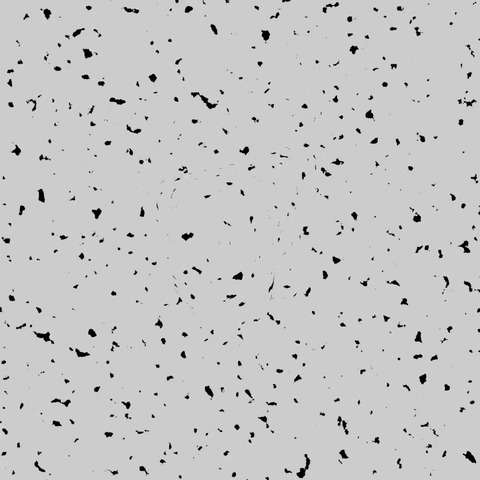} \\[0.15cm]
    \normalsize Texture 22 & \normalsize Texture 25
    \end{tabular}
    \caption{\textbf{Example Gestalt Stimuli with Different Textures.} First frame of scene 00000 rendered with seven different texture patterns. The Gestalt experiment uses 20 scenes (00000--00019), each rendered with these seven textures (00, 07, 13, 16, 21, 22, 25), yielding 140 total stimuli to evaluate structure-from-motion segmentation. In the paper, these textures are referenced sequentially as (00, 01, 02, 03, 04, 05, 06).}
\label{fig:gestalt_textures}
\end{figure*}

Figure~\ref{fig:gestalt_textures} shows an example of the Gestalt structure-from-motion stimuli with different textures. We visualize the first frame of scene 00000 rendered with seven different texture patterns. The Gestalt experiment uses 20 scenes (00000--00019), each rendered with these seven textures (00, 07, 13, 16, 21, 22, 25), yielding 140 total stimuli to evaluate structure-from-motion segmentation across diverse visual appearances.

\section{RGB 3D Inference}
\label{app:3d_first_frames}

\begin{table}[ht]
\centering
\caption{\textbf{Jaccard Index on Supplementary Videos.} We report the Jaccard index for GenMatter and CoTracker3 on each supplementary video. Best per video is bolded.}
\label{tab:supplementary_jaccard}
\begin{tabular}{lcc}
\toprule
\textbf{Video} & \textbf{GenMatter} & \textbf{CoTracker3} \\
\midrule
\texttt{cloth\_bag} & \textbf{0.84} & 0.32 \\
\texttt{gray\_jacket} & 0.91 & \textbf{0.98} \\
\texttt{jello} & \textbf{0.93} & 0.57 \\
\texttt{manta\_ray} & 0.96 & \textbf{1.00} \\
\texttt{eagle} & 0.86 & \textbf{0.89} \\
\texttt{ostrich} & 0.91 & \textbf{0.97} \\
\texttt{purple\_jacket} & 0.81 & \textbf{0.99} \\
\texttt{snake} & \textbf{0.93} & 0.47 \\
\texttt{whiskey\_swirl} & 0.49 & \textbf{0.79} \\
\texttt{wine\_swirl} & 0.67 & \textbf{0.97} \\
\bottomrule
\end{tabular}
\end{table}

\subsection{Experimental Details}

\paragraph{GenMatter Setup}

We provide additional technical implementation details for our TAP-Vid-DAVIS experiments.

\noindent
\textbf{Initialization} At frame 0, we perform 30 Gibbs sweeps to initialize all model parameters before sequential tracking begins. Particles are initialized through hierarchical K-means clustering on 3D positions lifted from tracked points using monocular depth estimates. For each particle, DINO features $\mathbf{f}_\ell^\mathcal{B}$ are initialized by averaging DINO descriptors over all pixels assigned to that particle. When SAM2 frame-0 masks are available, we adaptively determine the number of clusters $K$ based on the number of components in the mask rather than fixing $K$. We sample tracked points uniformly across each frame for the whole video.

\noindent
\textbf{Per-frame Gibbs schedule} For each frame $t > 0$, we apply a fixed schedule of blocked Gibbs updates. We apply updates to:
\begin{itemize}
    \item cluster-level rigid transformations $(\mathbf{R}_k, \mathbf{t}_k)$
    \item data point-to-particle assignments $z_n^\mathcal{B}$, conditioned on position likelihood only (with outliers disabled)
    \item data point-to-particle assignments $z_n^\mathcal{B}$, with full position-velocity-feature likelihood and $p_{\text{outlier}} = 0.1$
    \item particle spatial means $\boldsymbol{\mu}_\ell^\mathcal{B}$ and velocity parameters $(\mathbf{v}_\ell^\mathcal{B}, \boldsymbol{\Sigma}_\ell^{\mathcal{V}})$
    \item particle DINO features $\mathbf{f}_\ell^\mathcal{B}$
\end{itemize}

Multiple iterations are performed for spatial and velocity updates to ensure convergence. Mixture weights $\boldsymbol{\pi}^\mathcal{B}$ and $\boldsymbol{\pi}^\mathcal{H}$ are updated via Dirichlet conditionals after their respective assignment steps.

\noindent
\textbf{Outlier handling} During tracking (frames $t > 0$), we enable outliers by including an additional mixture component with weight $p_{\text{outlier}} = 0.1$. The outlier likelihood for a data point with velocity $\mathbf{v}_n$ is modeled as a Gamma distribution on speed $\|\mathbf{v}_n\|$ with shape parameter $\alpha$ and rate parameter $\beta$, which accounts for velocity outliers typically arising from unreliable motion estimates at object boundaries.

\paragraph{CoTracker3 Setup}

We run CoTracker3 with its default PyTorch Hub offline mode implementation. We initialize 500 query points in frame 0, matching the particle count used in GenMatter for fair comparison. Query points are randomly sampled uniformly across the first frame. Because we do not use ground-truth segmentation masks during initialization, query points are distributed uniformly across the object and background regions. As a result, the tracker cannot concentrate query points on the object, making it difficult to share statistical strength across object particles. We evaluate tracking quality using the same particle-based metrics as GenMatter, with the difference that at each frame, we only consider points which CoTracker3 has identified as visible. We first classify particles as object or background based on their frame-0 location relative to the segmentation mask, and we compute per-frame Jaccard by projecting tracked locations onto ground-truth masks at each timestep. The per-frame Jaccard indices are averaged to obtain the per-video Jaccard index.

\subsection{Supplementary Video Descriptions}
\begin{figure*}[p]
    \centering
    \begin{subfigure}[t]{0.85\textwidth}
    \includegraphics[width=\linewidth]{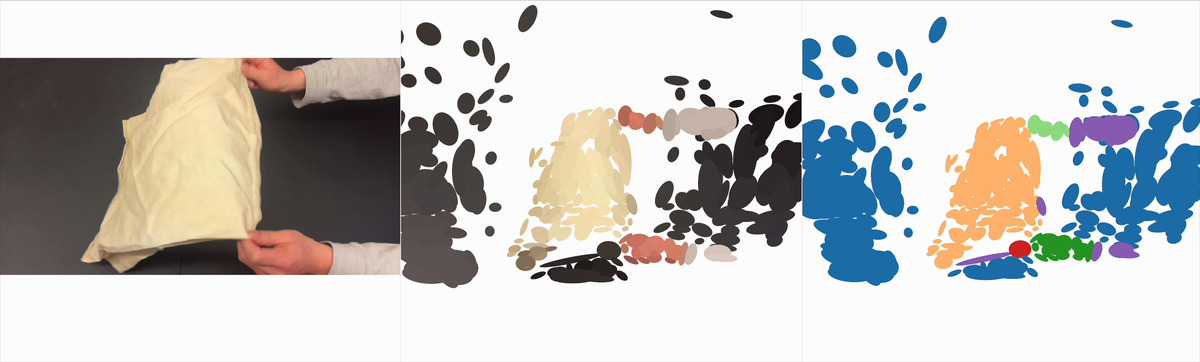}
        \caption{\texttt{cloth\_bag}}
        \label{fig:cloth_bag_first}
    \end{subfigure}

    \vspace{0.5em}

    \begin{subfigure}[t]{0.85\textwidth}
    \includegraphics[width=\linewidth]{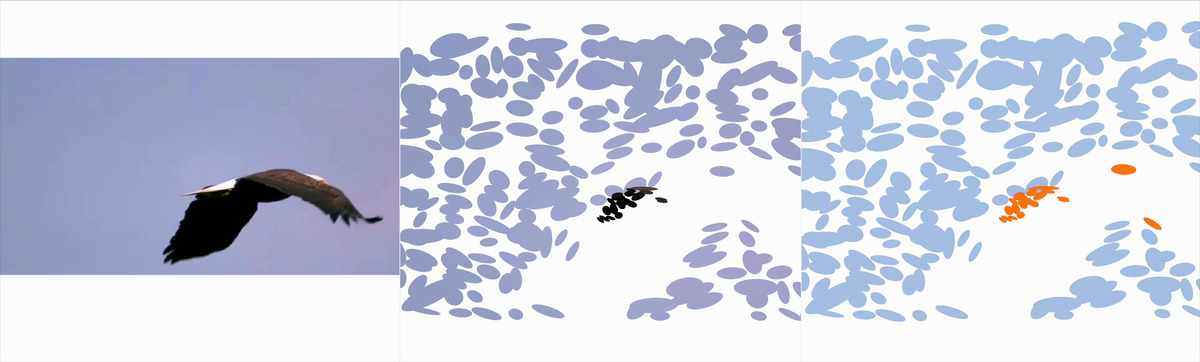}
        \caption{\texttt{eagle}}
        \label{fig:eagle_first}
    \end{subfigure}

    \vspace{0.5em}

    \begin{subfigure}[t]{0.85\textwidth}
    \includegraphics[width=\linewidth]{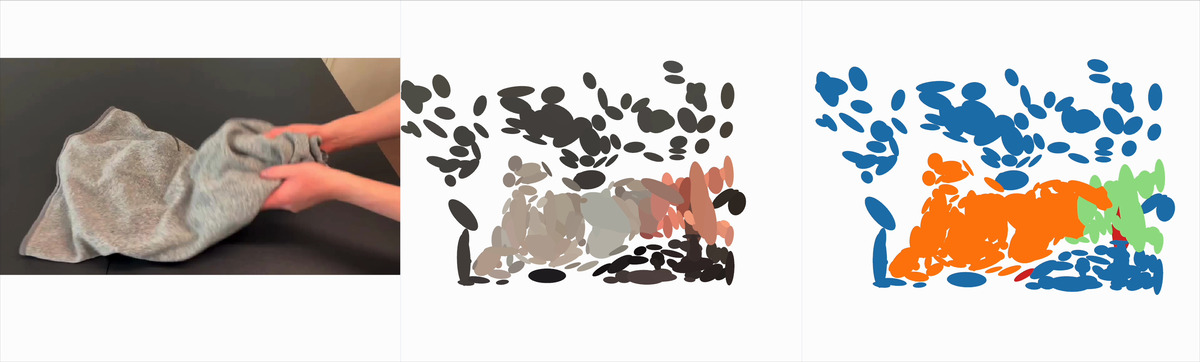}
        \caption{\texttt{gray\_jacket}}
        \label{fig:gray_jacket_first}
    \end{subfigure}

    \vspace{0.5em}

    \begin{subfigure}[t]{0.85\textwidth}
    \includegraphics[width=\linewidth]{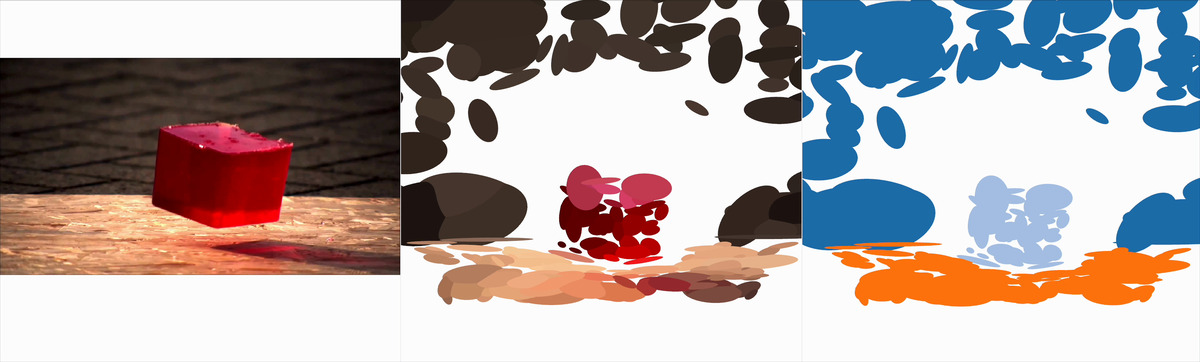}
        \caption{\texttt{jello}}
        \label{fig:jello_first}
    \end{subfigure}

    \caption{First frame visualizations of RGB 3D inference (Part 1). Each image shows the initial particle distribution and segmentation for the respective sequence.}
    \label{fig:3d_first_frames_1}
\end{figure*}

\begin{figure*}[p]
    \centering
    \begin{subfigure}[t]{0.85\textwidth}
    \includegraphics[width=\linewidth]{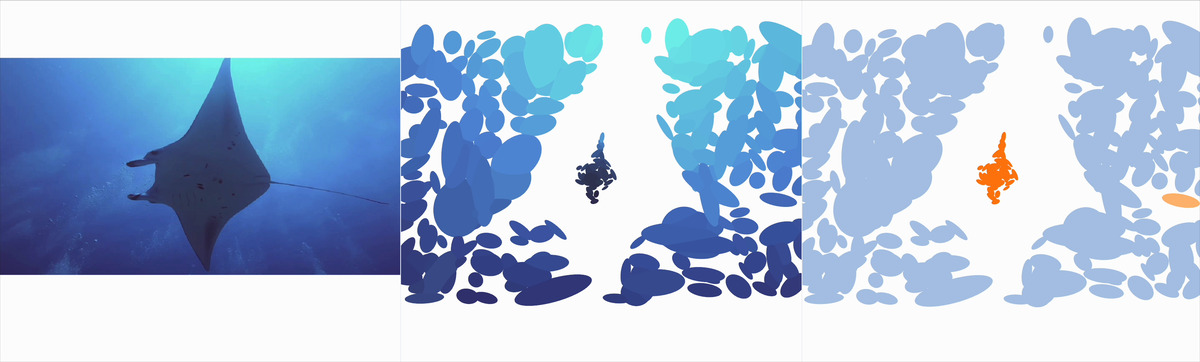}
        \caption{\texttt{manta\_ray}}
        \label{fig:manta_ray_first}
    \end{subfigure}

    \vspace{0.5em}

    \begin{subfigure}[t]{0.85\textwidth}
    \includegraphics[width=\linewidth]{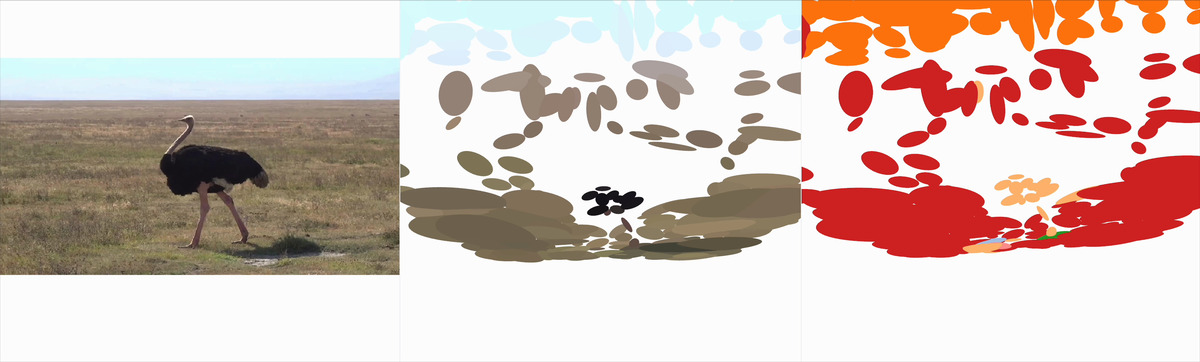}
        \caption{\texttt{ostrich}}
        \label{fig:ostrich_first}
    \end{subfigure}

    \vspace{0.5em}

    \begin{subfigure}[t]{0.85\textwidth}
    \includegraphics[width=\linewidth]{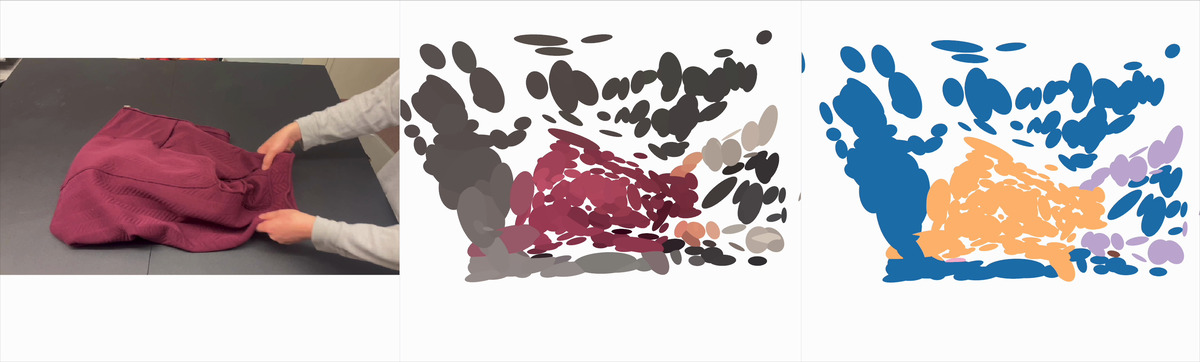}
        \caption{\texttt{purple\_jacket}}
        \label{fig:purple_jacket_first}
    \end{subfigure}

    \caption{First frame visualizations of RGB 3D inference (Part 2). Each image shows the initial particle distribution and segmentation for the respective sequence.}
    \label{fig:3d_first_frames_2}
\end{figure*}

\begin{figure*}[p]
    \centering
    \begin{subfigure}[t]{0.85\textwidth}
    \includegraphics[width=\linewidth]{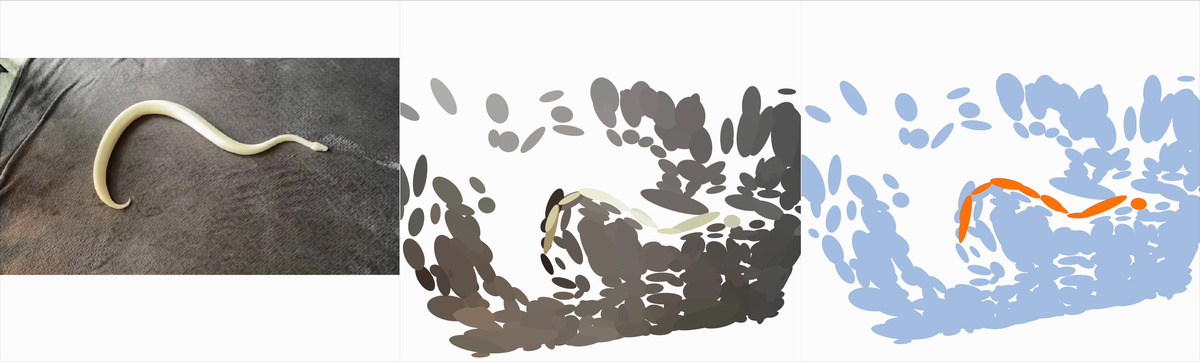}
        \caption{\texttt{snake}}
        \label{fig:snake_first}
    \end{subfigure}

    \vspace{0.5em}

    \begin{subfigure}[t]{0.85\textwidth}
    \includegraphics[width=\linewidth]{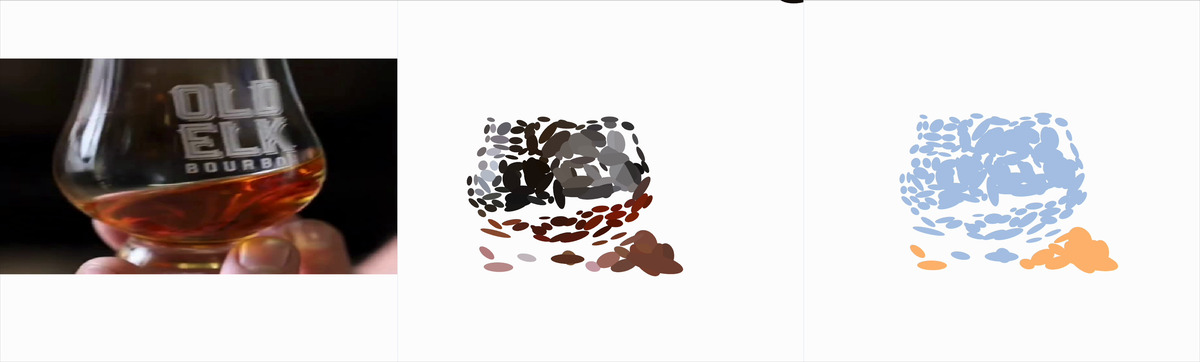}
        \caption{\texttt{whiskey\_swirl}}
        \label{fig:whiskey_swirl_first}
    \end{subfigure}

    \vspace{0.5em}

    \begin{subfigure}[t]{0.85\textwidth}
    \includegraphics[width=\linewidth]{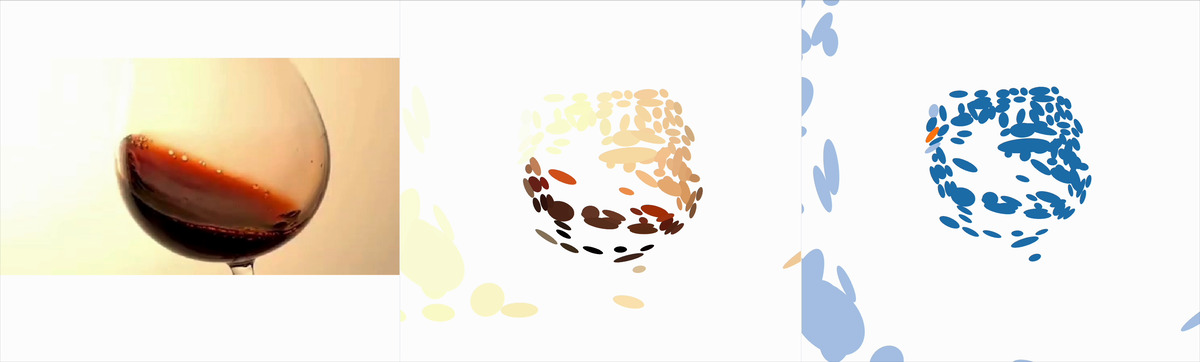}
        \caption{\texttt{wine\_swirl}}
        \label{fig:wine_swirl_first}
    \end{subfigure}

    \caption{First frame visualizations of RGB 3D inference (Part 3). Each image shows the initial particle distribution and segmentation for the respective sequence.}
    \label{fig:3d_first_frames_3}
\end{figure*}

We include supplementary videos that visualize the full particle-based inference process across time for the small qualitative deformable dataset introduced in the main paper: \texttt{cloth\_bag}, \texttt{gray\_jacket}, \texttt{jello}, \texttt{manta\_ray}, \texttt{eagle}, \texttt{ostrich}, \texttt{purple\_jacket}, \texttt{snake}, \texttt{whiskey\_swirl}, and \texttt{wine\_swirl}. These sequences span a range stuff and things observable in the physical world, including articulated structures, highly deformable solids, and liquids, allowing us to evaluate model performance across the full spectrum of matter interpretable by human vision. The first frame of particle tracking in these videos is shown in Figure ~\ref{fig:3d_first_frames_1} , Figure~\ref{fig:3d_first_frames_2}, and Figure~\ref{fig:3d_first_frames_3}.

\paragraph{Evaluation} GenMatter achieves higher average Jaccard (0.83 vs 0.79) on the small qualitative deformable dataset, with strong performance on highly deformable solid matter (\texttt{cloth\_bag}, \texttt{snake}, and \texttt{jello} in particular have highly deformable solid matter). It performs weakly on liquid (\texttt{wine\_swirl} and \texttt{whiskey\_swirl}) because the appearance of liquid makes it difficult to estimate matter motion, and liquid is particularly unstructured. On the other videos in the set, the performance of both models is similar. This pattern suggests GenMatter's probabilistic particle representation describes highly deformable solid matter better than it describes persistent liquid. The reported Jaccard indices in Table~\ref{tab:supplementary_jaccard} use SAM-generated masks as pseudo-ground-truth, as we do not have ground truth segmentation for these videos. Because GenMatter's initial particle clustering also uses SAM, this evaluation is not as robust as datasets derived from DAVIS. However, visual inspection confirms the SAM-generated masks are accurate, and this evaluation helps us bridge the gap towards more precise evaluation of 3D matter representations.

\paragraph{Visualization} The supplementary videos apply weight thresholding to particles before rendering. Many particles explain negligible data and have near-zero weights. Our model places little belief in the matter represented by these particles. Removing these low-weight particles improves visual clarity. However, hard thresholding causes flicker at the threshold boundary. Marginal particles will flicker across frames depending on whether their weight exceeds the cutoff. This flicker is a visualization artifact and not model instability.

{\small